\documentclass[10pt,twocolumn,letterpaper]{article}

\usepackage[pagenumbers]{cvpr} 
\usepackage{algorithm}
\usepackage{algorithmic}
\usepackage[dvipsnames]{xcolor}

\definecolor{cvprblue}{rgb}{0.21,0.49,0.74}
\usepackage[pagebackref,breaklinks,colorlinks,citecolor=cvprblue]{hyperref}

\def\paperID{3068} 
\def\confName{CVPR}
\def\confYear{2024}

\title{Merging Vision Transformers from Different Tasks and Domains}

\author{
Peng Ye\textsuperscript{\rm 1}$^\dagger$, Chenyu Huang\textsuperscript{\rm 2}$^\dagger$, Mingzhu Shen\textsuperscript{\rm 3}, \\
Tao Chen\textsuperscript{\rm 1}\thanks{Corresponding author.~~~$^\dagger$Equal Contribution}, Yongqi Huang\textsuperscript{\rm 1}, Yuning Zhang\textsuperscript{\rm 2}, Wanli Ouyang\textsuperscript{\rm 4}\\
\textsuperscript{\rm 1}Fudan University, \textsuperscript{\rm 2}Southeast University, \textsuperscript{\rm 3}Imperial College London, \textsuperscript{\rm 4}Shanghai AI Laboratory
}

\begin{document}
\maketitle
\begin{abstract}
This work targets to merge various Vision Transformers (ViTs) trained on different tasks (i.e., datasets with different object categories) or domains (i.e., datasets with the same categories but different environments) into one unified model, yielding still good performance on each task or domain. Previous model merging works focus on either CNNs or NLP models, leaving the ViTs merging research untouched. To fill this gap, we first explore and find that existing model merging methods cannot well handle the merging of the whole ViT models and still have improvement space.
To enable the merging of the whole ViT, we propose a simple-but-effective gating network that can both merge all kinds of layers (e.g., Embedding, Norm, Attention, and MLP) and select the suitable classifier. Specifically, the gating network is trained by unlabeled datasets from all the tasks (domains), and predicts the probability of which task (domain) the input belongs to for merging the models during inference.
To further boost the performance of the merged model,  especially when the difficulty of merging tasks increases,
we design a novel metric of model weight similarity, and utilize it to realize controllable and combined weight merging.
Comprehensive experiments on kinds of newly established benchmarks, validate the superiority of the proposed ViT merging framework for different tasks and domains. Our method can even merge beyond 10 ViT models from different vision tasks with a negligible effect on the performance of each task. 
%
\end{abstract}    
\section{Introduction}
\label{sec:intro}


With the progress of deep learning in different application fields~\cite{he2016deep,dosovitskiy2020image,ainsworth2022git, jin2022dataless,stoica2023zipit}, the number of pretrained or finetuned models grows explosively. This necessitates research on how to exploit these existing model weights increasingly essential.
As one critical technology, 
the model merging that combines the weights of multiple models into a single one has attracted increasing attention.
This technology can broaden the scope of the model's capabilities without additional training, thus showing broad application prospects.

\begin{figure}[t]
  \centering
   \includegraphics[width=0.95\linewidth]{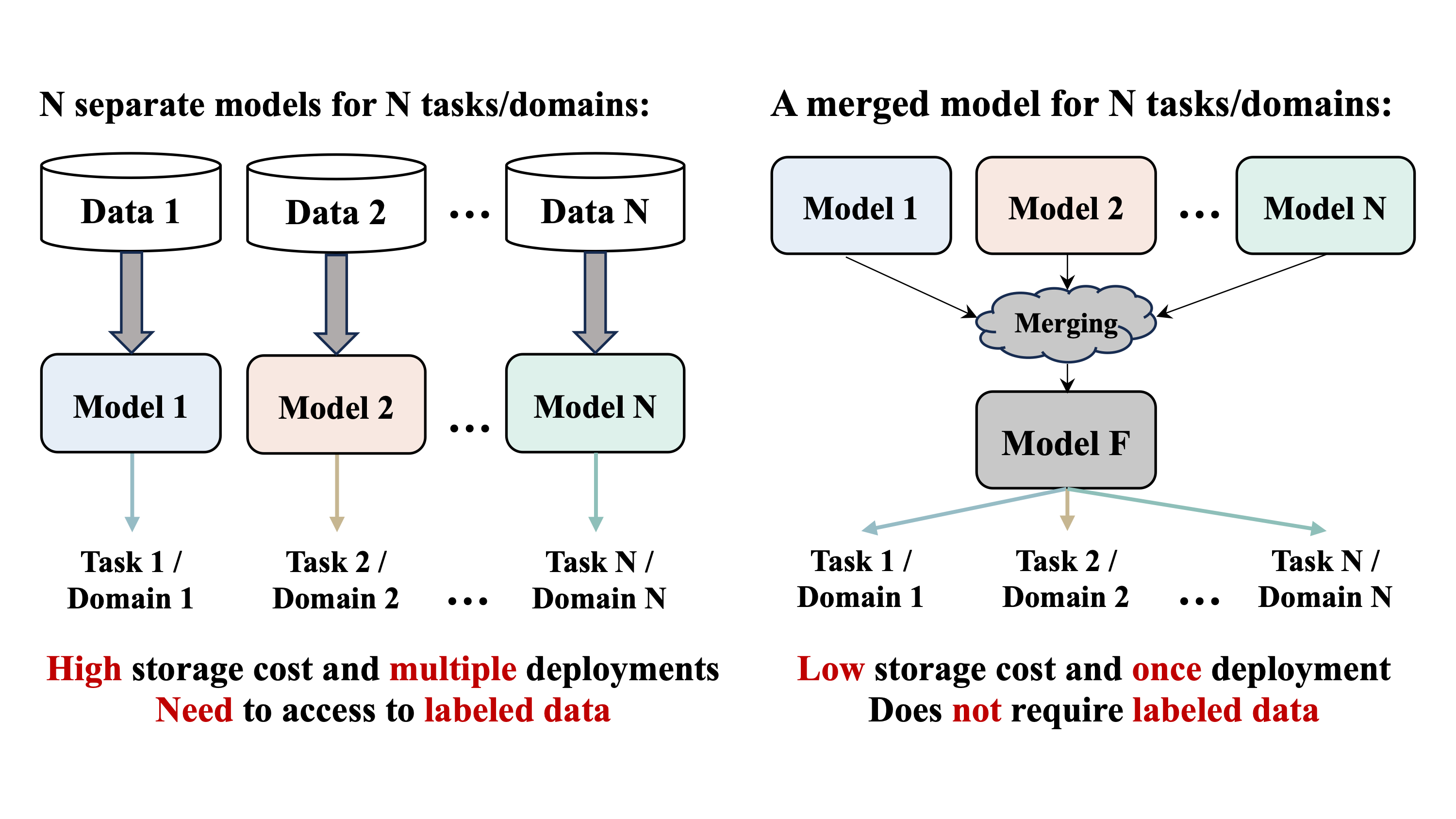}
   \vspace{-6mm}
   \caption{Merging models from different tasks (domains) can yield a single model that can handle N tasks or domains simultaneously. This will substantially reduce the storage and deployment costs and does not require access to labeled data.}
   \label{fig:motivation}
   \vspace{-0.1cm}
\end{figure}

\begin{table}[t]
\vspace{-1mm}
\caption{Comparisons among different model merging methods.}
\label{tab:previous_methods}
\vspace{-0.2cm}
\centering
\scalebox{0.68}{
\begin{tabular}{@{}lccccc@{}}
\toprule
& \begin{tabular}[c]{@{}c@{}} Different \\ Tasks \end{tabular} 
& \begin{tabular}[c]{@{}c@{}} Different \\ Domains
\end{tabular}
& \begin{tabular}[c]{@{}c@{}}Beyond \\ 10 Models \end{tabular}
& \begin{tabular}[c]{@{}c@{}}Model \\ Architecture \end{tabular} 
& \begin{tabular}[c]{@{}c@{}}Area of \\ Application \end{tabular} \\ 

\midrule

EMA &\texttimes &  \texttimes & $\checkmark$ & Agnostic & Agnostic \\

Model Soup &  \texttimes&  \texttimes & $\checkmark$ & Agnostic & Agnostic \\

ZipIt! & $\checkmark$ &   \texttimes &  \texttimes  & CNN & Vision\\

RegMean & $\checkmark$ &  $\checkmark $&  \texttimes & Attn$\And$MLP & NLP\\

MMM & $\checkmark$ &   \texttimes&  \texttimes & Attn$\And$MLP & Multi-modal\\

\textbf{Ours} & $\checkmark$ & $\checkmark$ & $\checkmark$ & Whole ViT & Vision\\
\bottomrule
\end{tabular}
}
\vspace{-0.2cm}
\end{table}

Early studies, including Exponential Moving Average (EMA)~\cite{wightman2021resnet,cai2021exponential}, Model Soup~\cite{wortsman2022model}, and DiWA~\cite{rame2022diverse}, have illustrated the effectiveness of combining weights from multiple rounds of updates within the same task. It has also been demonstrated by other studies like Model Ratatouille~\cite{rame2023model} and Fisher-weighted averaging~\cite{matena2022merging} that merging weights from different tasks can improve a specific single-task performance. To further expand an existing model’s capabilities, the latest model merging method broadens the settings of these methods and tries to merge multiple models from different tasks or domains to obtain a single model performing well on all the tasks or domains. Compared with separate models, which require high storage cost, multiple deployments, and access to labeled datasets, the merged model shows the advantages of low storage cost, once deployment, and needs no labeled data. The detailed comparisons are shown in~Fig.~\ref{fig:motivation}.

Recently, a few works have attempted this challenging but significant task.
RegMean~\cite{jin2022dataless} proposes an effective solution for merging NLP models fine-tuned on different domains and tasks by giving a new closed-form solution to the merging problem.
MMM~\cite{sung2023empirical} verifies the effectiveness of existing methods for merging transformers trained on different modalities.
However, both methods cannot handle the whole ViT model. For example, RegMean can only handle attention and MLP layers because RegMean only works when the dimensions of the weights and the input features are matched,
and both methods can not work for classifiers.
ZipIt~\cite{stoica2023zipit} focuses on merging models trained on disjoint vision tasks by identifying the tasks' shared features on convolutional neural networks (CNNs) instead of transformer-based models.
Unfortunately, all these methods suffer significant performance drops when the merging task becomes more difficult, i.e. merging more models or merging models on more challenging tasks (domains).
More importantly, as shown in the comparisons of representative merging methods in Tab.~\ref{tab:previous_methods}, there is currently no research or study available on how to merge the whole Vision Transformers (ViTs) from different tasks (domains).

To fill this research gap, we first explore the effectiveness of merging ViTs using existing methods with different weights initialization strategies. Our empirical studies show that existing methods demonstrate varying degrees of effectiveness when merging ViTs from different tasks (domains) provided that they are from the same pertraining, otherwise all these methods easily fail.
In addition, some unsolved problems with the existing methods are discovered including not applicable to partial layers and significant performance decay. 
Please refer to~\cref{sec:Empirical Studies} for more details. We believe that the performance of merging the ViT models and its applicability can be greatly improved by solving these problems.

In this paper, we propose a simple-but-effective gating network with multiple functions, including merging all kinds of layers within a typical ViT (e.g., Embedding, Norm, Attention, and MLP) and automatically selecting the proper classifier.
Specifically, the gating network is trained using unlabeled datasets from all the tasks (domains), and used to predict the probability distribution of the input image belonging to each task or domain, which is then utilized to merge layers and select the classifier during inference.
To further boost the performance of the merged model, especially when the difficulty of merging tasks increases,
a novel metric of weight similarity is designed and utilized to realize controllable and combined weight merging for different layers. Comprehensive experiments on kinds of newly-built benchmarks validate the superiority of our method, which can successfully merge up to 12 models with negligible performance drops. Our contributions are summarized as: 
\begin{itemize}
    \item We investigate the problem of how to merge ViTs from different tasks (domains) for the first time. After validating the applicability of existing CNN and NLP merging methods for ViTs merging, we identify some insightful observations and their non-neglectable problems.

    \item We propose a new gating-based model merging method, which can both merge kinds of layers and automatically select the proper classifier, enabling the merging of the whole ViT. The gating network is trained using only unlabeled datasets from different tasks (domains), without additional data collection and labeling costs.
    

    
    \item We design a novel metric of weight similarity to adaptively select partial layers to be handled by the gating network, which can realize controllable and combined weight merging for different layers.
    
    \item We validate the effectiveness of the proposed method by comprehensive experiments on kinds of newly-established benchmarks. Our method is the first to successfully merge beyond 10 ViT models from different vision tasks with a negligible effect on performance.
\end{itemize}
\section{Related Works}
\label{sec:related_work}

\textbf{Model Merging} that fuses the weights of multiple models into one single model is receiving increasing attention. Various merging methods and applications have been proposed. Early research focuses on fusing weights trained or finetuned on the same task to improve the performance, enhance the generalization ability, or provide a better teacher, such as Model soup~\cite{wortsman2022model}, DiWA~\cite{rame2022diverse}, and EMA~\cite{wightman2021resnet,cai2021exponential}.
Besides, although Model Ratatouille~\cite{rame2023model} and Fisher-weighted averaging~\cite{matena2022merging} attempt to merge model weights from different tasks, they focus on improving the single-task performance. Recently, RegMean~\cite{jin2022dataless} merges model weights by computing a closed-form solution for each individual attention or MLP layer, and verifies its effectiveness when merging different tasks (domains) in NLP. MMM~\cite{sung2023empirical} empirically studies existing merging methods for multimodal models. Both methods only merge part of the transformer model. ZipIt~\cite{stoica2023zipit} proposes the first work to merge the CNN models trained on disjoint vision tasks. However, ZipIt focuses on merging CNNs instead of transformer-based models.
Besides, all the above methods suffer significant performance drops when the difficulty of merging tasks increases. In this paper, we explore how to merge multiple ViTs from different tasks (domains) into a single model that can perform well on all tasks (domains) for the first time. 

\textbf{Multi-Task Learning} trains a single model via the joint datasets of multiple tasks, allowing the model to leverage information across different tasks~\cite{vandenhende2021multi, zhang2023uni3d, zhang2021survey}. In general, multi-task learning typically necessitates access to the labeled data of multiple tasks for training the model from scratch. As the number of different tasks increases, the optimization process becomes more challenging~\cite{vandenhende2021multi, zhang2021survey}. In comparison, our method eliminates the need for labeled data and solely relies on released model weights, which is particularly practical in scenarios with limited computing resources and where data privacy is a priority. Moreover, our method has the capability to merge beyond 10 ViT models from different tasks.

\textbf{Multi-Domain Generalization} can be roughly divided into domain adaption and generalization methods. Domain adaption enables the model learned from the source domain to perform well on the target domain through various strategies such as feature alignment~\cite{mei2023denkd}, distribution matching~\cite{mei2023automatic}, and reweighting of samples~\cite{yuan2023bi3d}. Domain generalization enhances the model's generalization ability for different (even unseen) domains through techniques such as domain augmentation~\cite{zhang2023resimad} and regularization methods~\cite{zhou2022domain}. Differently, our method enables a single model to perform well on multiple domains by merging the ViT models from different domains, which is orthogonal and complementary to existing domain adaption or generalization works.


\begin{figure*}[t]
  \centering
   \includegraphics[width=0.95\linewidth]{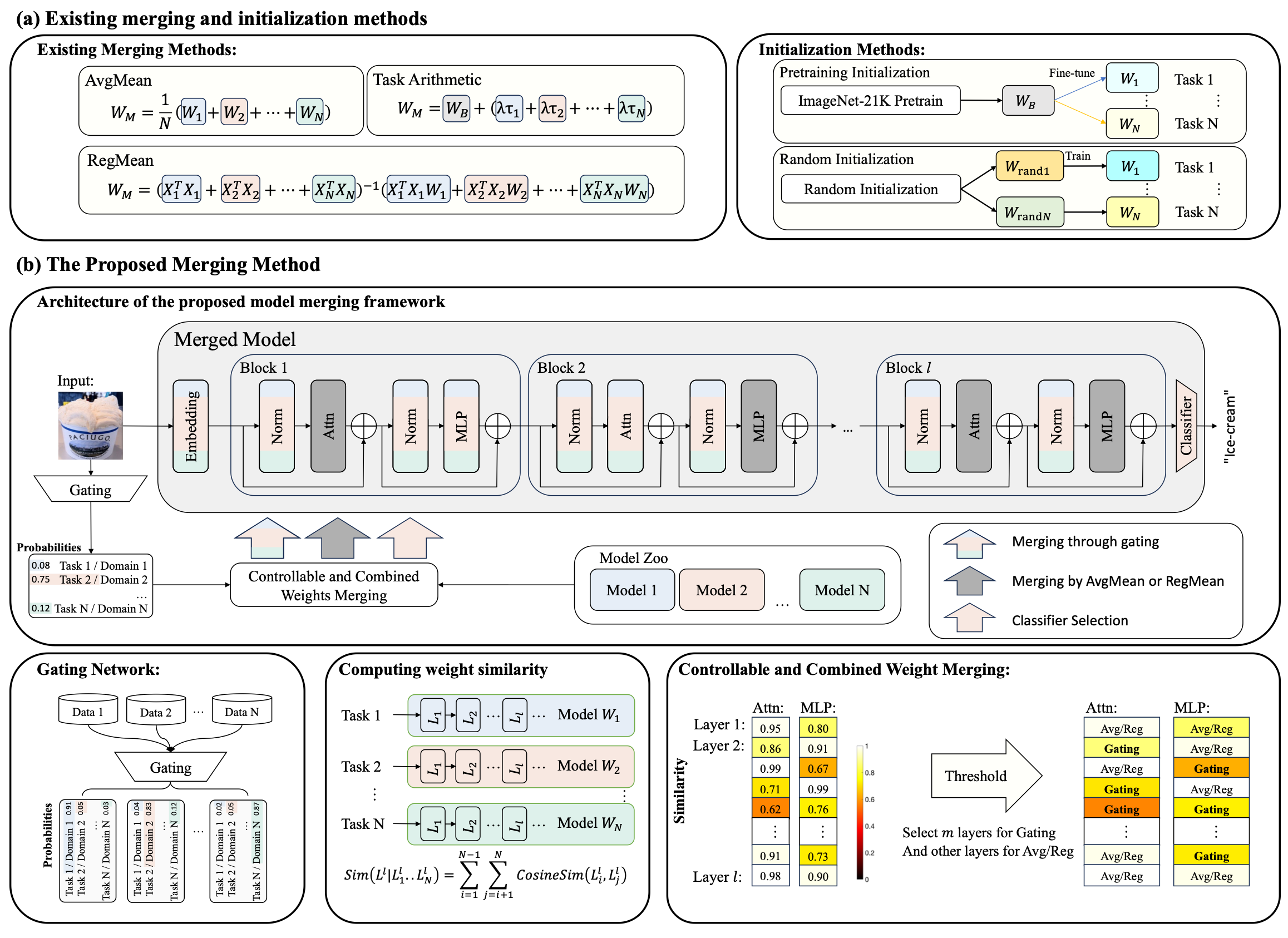}
   \vspace{-3mm}
   \caption{Since merging ViTs has not been investigated before, we first explore the effectiveness of (a) existing merging methods with different initialization methods. Then, we present (b) the proposed merging method. To enable the merging of the whole ViT models, we propose a simple-but-effective gating network trained by unlabeled datasets, which can merge all kinds of layers (e.g., Embedding, Norm, Attention, and MLP) and select the classifier.
   To further boost the performance of the merged model, we design a novel metric of weight similarity, and utilize it to realize controllable and combined weight merging.
    }

   \label{fig:onecol}
   \vspace{-0.2cm}
\end{figure*}

\section{Methods}
\label{sec:methods}

We aim to obtain a single model applicable to multiple tasks (domains) by merging the model weights specifically fine-tuned for these tasks (domains). In this section, we start with introducing and comparing some of the existing model merging methods, followed by elaborating on the proposed method to solve the problems of existing merging methods and improve the merged model's performance. Detailed information about the algorithm flow of our method can be found in Appendix B.

\subsection{Empirical Studies, Observations, and Problems} \label{sec:Empirical Studies}
First, we briefly introduce three existing model merging methods shown in~\cref{fig:onecol}(a). 
\textbf{Simple Average Mean (AvgMean)} element-wise averages the weights of all the models to be merged. Though simple, its effectiveness has been demonstrated~\cite{rame2023model, wortsman2022model, sung2023empirical}.
\textbf{Task Arithmetic} edits models using task vector, which denotes the difference between the specifically fine-tuned weights and the initial weights~\cite{ilharco2022editing}. By adding task vectors together, a multi-task model can be built~\cite{ilharco2022editing}.
\textbf{RegMean} merges linear layers by finding closed-form solutions for the weights from the same initialization~\cite{jin2022dataless}. 
For detailed information about these methods, please check Appendix A.

As merging ViTs from different tasks (domains) has not been investigated before, we first conduct empirical studies to compare existing merging methods with or without the pretraining initialization, as shown in Tab.~\ref{tab:merge_3_and_6_models} and~\ref{tab:merge_12_models}. Based on these results, we give two \textbf{insightful observations}:
1) Existing merging methods show varying degrees of effectiveness for merging ViTs from different tasks (domains). Specifically, when an appropriate $\lambda$ is selected, task arithmetic can perform better than AvgMean. Moreover, though proposed for solving NLP tasks, RegMean demonstrates its superiority by exceeding both AvgMean and task arithmetic. 2) All the above merging methods easily fail when the model weights are trained from scratch independently, and they exhibit notable improvements when the model weights are sourced from the same pre-training. This may be attributed to the linear mode connectivity (LMC), which claims that models sharing a common initial optimization path are linearly connected with a low loss barrier \cite{jain2023dart, gotmare2018using, rame2023model}. That is to say, models from the same pre-training belong to a common basin and thus may benefit from weights merging. Generally, it is easy to meet such conditions under the current prevailing research paradigm of pretraining then fine-tuning.

Though these merging methods can be applied to vision tasks, there are some \textbf{unsolved problems}:
1) All merging methods mentioned above cannot handle classifiers. For one thing, the output dimensions of classifiers for different tasks are not necessarily the same. For another, even if dealing with multi-domain tasks where the classifiers' dimensions are the same, merging the classifier may have a negative impact on performance~\cite{jin2022dataless,rame2023model}.
Thus, existing methods usually assume that we know which task the input belongs to and select the classifier from the corresponding model, which is impractical in applications.
2) Though showing better performance than other merging methods, RegMean only merges the weights of attention and MLP layers within the transformer blocks, and other weights, such as Norm, Embedding, and bias items, are merged by AvgMean, where we believe there is still some room for optimization.
3) All the afore-mentioned merging methods suffer significant performance drops when the difficulty of merging tasks increases. i.e. merging a larger number of tasks or tasks to be merged vary greatly, as we demonstrate in Tab.~\ref{tab:merge_3_and_6_models},~\ref{tab:merge_12_models} and~\ref{tab:merge_3_models_imnet}.

Therefore, we propose a model merging method to solve the existing problems and promote the practical application of model merging. The structure of the method is shown in Fig.~\ref{fig:onecol}~(b). The details are shown below.

\subsection{Multi-Task Gating Network For Both Kinds of Weights Merging and Classifier Selection} \label{sec:gating}
We hope to obtain a merged model $f_M$ using $N$ ($N \geq 2$) models $[f_1...f_N]$ where $f_k=[W_k^1...W_k^l, C_k], k \in [1..N]$, and some of their unlabeled data $[X_1...X_N]$ from the test or validation set. All models are from the same initialization and fine-tuned on different tasks (domains) $[T_1...T_N]$. This setting is consistent with RegMean~\cite{jin2022dataless}, which needs unlabeled data to compute gram matrices. To solve the problems mentioned above, we then introduce a simple-but-effective gating network, which can both select the appropriate classifier and merge kinds of layers, as shown in Fig.~\ref{fig:onecol}.

\textbf{Training Stage.} Before applied to model merging, the gating network needs to be trained in advance. The dataset used for training the gating network is composed of the unlabeled dataset for each task, and the label for each input is defined as its corresponding task number. Compared with common vision tasks, training the gating network is a rather simple task since we only need to judge the probabilities that the input is from each task.
Thus, a lightweight CNN, e.g. MobileNetV2~\cite{Sandler_2018_CVPR}, can handle this task without a significant increase in the parameter number when merging models. Besides, we find that finetuning the gating network on a pretrained model benefits fast convergence.

\textbf{Inference Stage.} The gating network $G$ takes $x$ as input, and outputs the probability $P\left( x \right)=[p_1...p_N]$ for kinds of layers to be merged such as attention, MLP, Norm, and Embedding layers,
as follows:  
\begin{equation}
P \left( x \right) = SoftMax \left( G \left( x \right) \right)
\end{equation}

The merging process is a weighted average of corresponding layers of models from different tasks (domains). For instance, the $l$-th merged layer $W_M^l$ is computed from the corresponding layers of $N$ models $[W_1^l...W_N^l]$ as:
\begin{equation}
W_M^l = \sum_{i=1}^{N} p_i \cdot W_i^l
\end{equation}

Meanwhile, the gating network is used to choose rather than merge the classifier for the mentioned problem. Based on the output probability, the classifier of the merged model $C_M$ can be selected from $N$ classifiers $[C_1...C_N]$ as:
\begin{equation}
\begin{split}
C_M = C_i, \
i = \mathop{\mathrm{argmax}} \limits _{i=1..N} \left( P \left( x \right) \right)
\end{split}
\end{equation}

\begin{figure}[t]
  \centering
   \includegraphics[width=1.0\linewidth]{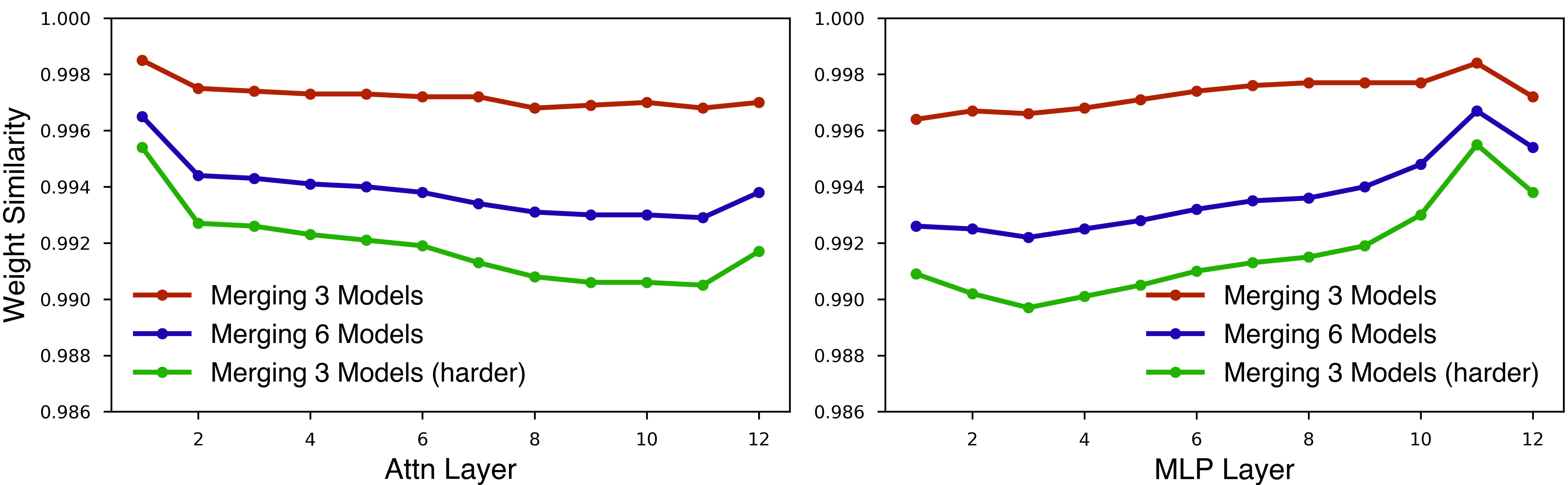}
   \caption{Weight similarity of each layer when merging 3 and 5 models. As a comparison, we merge another 3 models fine-tuned on harder tasks including ImageNet-1K~\cite{deng2009imagenet}.}
   \label{fig:weight_sim}
   \vspace{-0.2cm}
\end{figure}

\begin{table}[t]
\caption{Performance comparisons of the merged network with different merging strategies, verifying the superiority of Gating.}
\label{tab:gating_effectiveness}
\vspace{-0.2cm}
\centering
\scalebox{0.7}{
\begin{tabular}{@{}ccccc@{}}
\toprule
\begin{tabular}[c]{@{}c@{}} 
Attn $\And$ MLP \\ Embedding $\And$ Norm\\Classifier
\end{tabular}&
\begin{tabular}[c]{@{}c@{}} 
RegMean\\AvgMean\\Manual
\end{tabular}&
\begin{tabular}[c]{@{}c@{}} 
RegMean\\Manual\\Manual
\end{tabular}&
\begin{tabular}[c]{@{}c@{}} 
RegMean\\Gating\\Gating
\end{tabular}&
\begin{tabular}[c]{@{}c@{}} 
RegMean+Gating\\Gating\\Gating
\end{tabular}
\\
\cmidrule(r){1-1} \cmidrule(l){2-5}
Acc & 86.95 &88.30 & 88.41 & \textbf{93.40}
\\
\bottomrule
\end{tabular}
}
\end{table}

We further verify the effectiveness of gating based (task correlation prediction based) weights merging method. In detail, we merge 12 models using 4 different strategies: 1) The original RegMean settings, i.e. applying RegMean to attention and MLP layers, while applying AvgMean to  Embedding and Norm layers and manually selecting the classifier.
2) Applying RegMean to attention and MLP layers and manually selecting the Embedding and Normalization layers, and classifier from the model corresponding to the task the input comes from.
3) Applying RegMean to Attention and MLP layers and using the gating network to merge Embedding and Norm layers and automatically select the classifier.
4) Using the gating network to merge partial Attention and MLP layers, Embedding and Norm layers, and automatically select the classifier, and applying RegMean to the rest of attention and MLP layers.
The results are shown in \cref{tab:gating_effectiveness}. This validates the effectiveness of the gating network when applied to merge layers of each type in ViTs.

 \subsection{Multi-Task Weights Similarity for Controllable and Combined Weights Merging}
To improve the performance of the merged model and be able to merge models from more or more difficult tasks (domains), we further design a novel metric of weight similarity and use it to realize controllable and combined weight merging, as shown in Fig.~\ref{fig:onecol}.


\textbf{Multi-Task Weights Similarity.}  One of the biggest limitations to the performance of the merged model may come from the non-negligible differences among the model weights for different tasks (domains), which all existing merging methods cannot handle well. Therefore, we design a novel metric to measure the total weights similarity of each layer of models from different tasks.
In detail, the formula for calculating the total similarity of layer $l$ from all the $N$ model weights $[W_1...W_N]$ is as follows. 

\begin{equation} \label{eq:sim}
Sim\left(L^l|W_i^l..W_N^l\right)=\sum\limits_{i=1}^{N-1}\sum\limits_{j=i+1}^N CosineSim \left(W_i^l, W_j^l\right)
\end{equation}

\textbf{Controllable and Combined Weights Merging.} With the above metric, layers with high weights similarity can be merged through existing methods like AvgMean/RegMean, while other layers can be merged through our gating network. Specifically, the similarity of attention and MLP modules in each layer are computed separately. For the similarity of each attention module, all attention-related weights in a specific layer of a model are concatenated, and the final similarity is computed as the sum of the pairwise similarities of concatenated attention weights of different models. The similarity computation of each MLP module is similar.
Then, the similarities of all attention and MLP modules are stored separately, and $m$ attention layers and $m$ MLP with the lowest weights similarity are selected for gating-based merging while others are for AvgMean/RegMean-based merging.

Generally, a larger $m$ results in a larger number of parameters and higher performance of the merged model, and the selection of $m$ depends on the device's storage and the required performance of the merged model.
\cref{fig:weight_sim} shows the weight similarity among the corresponding attention and MLP layers when merging different models. As the difficulty of the merging task increases, i.e., merging more models or merging models fine-tuned on more challenging tasks, the weight similarity among the models decreases. Furthermore, different types of layers present different weight similarity distributions. For example, the attention layers generally show a decreasing trend in weight similarity, while the MLP layers show an increasing trend as the layer number increases. Therefore, we separately select the attention and MLP layers to be merged through gating.

In Sec.~\ref{sec:gating}, we demonstrate that
applying the gating network to merge layers in ViTs instead of AvgMean or RegMean can improve the merged model's performance. This is mainly because RegMean and AvgMean suffer significant performance decay when the model weights share a low similarity, while the proposed gating based (task correlation based) method can handle the situation better.

\begin{table*}[t]
\centering
\caption{Performance comparisons when merging ViT models from \textbf{3 or 6 tasks}. We first evaluate existing methods like AvgMean, Task Arithmetic, and RegMean, and further demonstrate our method can seamlessly integrate with these methods and enhance their performance.}
\label{tab:merge_3_and_6_models}
\vspace{-0.2cm}
\scalebox{0.56}{
\begin{tabular}{@{}lcccccccccccccccc@{}}
\toprule
 & \multicolumn{6}{c}{\textbf{Merging Models from 3 Different Tasks}} & \multicolumn{9}{c}{\textbf{Merging Models from 6 Different Tasks}} \\
\cmidrule(r){1-1} \cmidrule(lr){2-7} \cmidrule(l){8-16} 
Method &Params(M)&FLOPs(G)&MNIST&CIFAR-10 & EuroSAT & \textbf{Avg($\uparrow$)} & Params(M) & FLOPs&MNIST & CIFAR-10 & Vegetables & Food-101 &Kvasir-V2 & EuroSAT&\textbf{Avg($\uparrow$)} \\

\cmidrule(r){1-1} \cmidrule(lr){2-3} \cmidrule(lr){4-7} \cmidrule(lr){8-9}\cmidrule(l){10-16}  
Individual & 257.50& 35.14& 99.22 & 97.88 & 98.95 & 98.65 &515.07&35.14&  
99.22 & 97.88 & 100.00& 87.89 & 93.59&98.95& 96.26 \\

\cmidrule(r){1-7} \cmidrule(l){8-16} 

AvgMean(pretrained) & 85.85 &35.14&95.98 & 93.90 & 82.88 & 90.92 & 
85.94 & 35.14 & 68.92 & 78.91 & 96.66 & 78.45 & 46.08 & 55.71 & 70.79  \\
AvgMean(from-scratch)& 85.85 &35.14&13.20 & 14.74 & 8.67&12.20& 85.94 &35.14&6.75 & 10.02 & 6.82&5.93& 4.48& 7.29&6.88 \\

\cmidrule(r){1-7} \cmidrule(l){8-16} 
RegMean(pretrained) & 85.85 &35.14& 98.92 & 97.71 & 97.76 & 98.13 & 85.94&35.14& 98.41 & 96.73 & 97.63 & 85.41 & 80.91 & 94.16 & 92.21 \\
RegMean(from-scratch)& 85.85 &35.14&12.09&13.97&14.16&13.41&85.94&35.14&7.23&9.88&6.21 & 8.79 & 12.85& 7.31 &8.71\\

\cmidrule(r){1-7} \cmidrule(l){8-16} 

Task Arithmetic

& 85.85&35.14 &98.48 & 94.78 & 89.47 & 94.24 &
85.94&35.14&78.52 & 85.41 & 99.67 & 81.30 & 55.50 & 57.78 & 76.36 \\



\cmidrule(r){1-7} \cmidrule(l){8-16} 

Ours + AvgMean\\
$m = 0$ &  89.72 &35.78& 96.18 & 94.94 & 85.19 & 92.10 & 92.15&35.78
&68.49 &79.08 & 98.90 & 78.47 & 45.50 & 55.72 & 71.03 \\

$m = 1 $ &110.98 &35.78& 95.89 & 95.60 & 86.86 & 92.78 & 134.67&35.78& 64.19 & 81.54 & 99.06 & 77.42 & 49.67 & 58.05 & 71.66 \\

$m = 3$ & 153.50 &35.78& 96.28 & 96.75 & 92.87 & 95.30  &219.72&35.78& 87.87& 95.84 & 99.83 & 84.30 & 74.80 & 81.21 & 87.31 \\

$m = 5 $& 196.02 &35.78& 98.24 & 97.71 & 96.96 & 97.64  &304.76&35.78& 93.18&97.19 & 99.90 & 83.66 & 82.13 & 91.24 & 91.22 \\

\cmidrule(r){1-7} \cmidrule(l){8-16} 

Ours + RegMean\\
$m = 0$ &  89.72&35.78 &98.94 & 97.78 & 98.46 & 98.39& 92.15 &35.78 & 98.48 &96.98 & 97.93 & 85.81 & 85.25 & 95.92 & 93.40 \\

$m = 1$ & 110.98&35.78& 98.95 & 97.83 & 98.56 & 98.45  & 134.67&35.78&98.58 & 94.78 & 99.86 & 85.62 & 89.70 & 96.65 & 94.20 \\

$m = 3$  & 153.50&35.78& 99.02 & 97.83 & 98.67 & 98.51 &
219.72&35.78& 98.75 & 97.54 & 99.86 & 86.90 & 90.35 & 97.40 & 95.13 \\

$m = 5 $&  196.02&35.78& 99.09 & 97.86 & 98.78 & 98.58  & 304.76 & 35.78 & 98.95 & 97.80 & 99.90 & 86.98 & 92.57 & 98.13 & 95.72 \\

\bottomrule
\end{tabular}
}
\end{table*}

\begin{table*}[t]
\caption{Performance comparisons when merging ViT models from \textbf{12 tasks}.}
\label{tab:merge_12_models}
\vspace{-0.3cm}
\centering
\scalebox{0.6}{
\begin{tabular}{@{}lccccccccccccccc@{}}
\toprule
 & \multicolumn{15}{c}{\textbf{Merging Models from 12 Different Tasks}}\\
\cmidrule(r){1-1} \cmidrule(l){2-16}
Method &Params(M) &  FLOPs(G)&
MNIST & CIFAR-10 & Vegetables & Food-101 &Kvasir-v2& Cars & Intel Images & EuroSAT & Weather &Cats\&dogs & Mango & beans & \textbf{Avg($\uparrow$)}\\

\cmidrule(r){1-1}\cmidrule(lr){2-3} \cmidrule(l){4-16}

Individual&1030.20&35.14 &99.22 & 97.88 & 100 & 87.89 & 93.59 & 94.88 & 85.95 & 98.95 & 98.19 & 99.92 & 100 & 97.79&96.19\\

\midrule

AvgMean & 86.12&35.14&34.74 & 56.42 & 90.93 & 73.26 & 30.05 & 88.76 & 10.07 & 35.3 & 74.82 & 95.39 & 87.72 & 82.76&63.35\\

\midrule

RegMean &  86.12&35.14&97.69 & 94.69 & 98.13 & 81.73 & 72.22 & 93.65 & 35.24 & 89.47 & 90.75 & 99.04 & 99.27 & 91.57&86.95\\
\midrule
Task Arithmetic
&  86.12&35.14&48.94 & 65.67 & 93.16 & 74.34 & 35.47 & 90.56 & 10.60 & 31.45 & 79.07 & 96.27 & 92.55 & 84.29 &66.86\\

\midrule
Ours + AvgMean\\
$m = 0$&96.97&35.78&38.34 & 59.59 & 92.66 & 72.06 & 32.66 & 88.63 & 10.67 & 36.83 & 76.24 & 96.15 & 89.67 & 83.53 &64.75\\
$m = 1$&182.01&35.78&40.38 & 62.95 & 92.56 & 72.39 & 35.79 & 88.36 & 10.96 & 37.79 & 79.03 & 95.61 & 91.62 & 87.36 &66.23\\
$m = 3$&352.10&35.78&65.95&  88.7  & 97.9  & 78.55 & 53.24 & 92.53 & 30.71 & 63.84 & 89.93 & 98.63 & 98.67 & 91.95 &79.21\\
$m = 5$&522.19&35.78&85.49 & 96.83 & 99.83 & 81.99 & 77.57 & 93.46 & 52.67 & 85.73 & 94.16 & 99.14 & 99.70  & 95.79 &88.53\\

\midrule
Ours + RegMean\\
$m = 0$&96.97&35.78&97.80  & 95.94 & 98.16 & 81.89 & 78.86 & 93.56 & 39.77 & 92.68 & 91.71 & 98.93 & 99.62 & 91.95 &88.41\\

$m = 1$&182.01&35.78&97.95 & 96.04 & 98.26 & 81.75 & 84.15 & 93.73 & 42.89 & 94.09 & 92.90  & 98.96 & 99.80  & 91.95 &89.37\\

$m = 3$ & 352.10 & 35.78&98.65 &97.65 & 98.40 &85.09& 90.38 & 94.70 & 70.49 & 97.09 & 95.76 & 99.11 & 99.95 & 93.49 & 93.40\\

$m = 5$&522.19&35.78&98.98 & 97.97 & 98.53 & 86.22 & 92.52 & 94.40  & 79.95 & 98.06 & 96.88 & 99.15 & 99.95 & 95.02 &94.80\\

\bottomrule
\end{tabular}
}
\end{table*}

\section{Experiments and Results}
\label{sec:results}

\subsection{Experimental setup}
The main goal of our experiments is to benchmark the performance of different merging methods and compare them with individual models' performance before merging. 

\textbf{Datasets.} 
In order to fully evaluate the performance of the proposed method, our experiments involve 16 different  datasets, including
\textit{MNIST}~\cite{lecun1998mnist},
\textit{CIFAR-10}~\cite{krizhevsky2009learning},
\textit{Vegetables} \cite{ahmed2021dcnn},
\textit{Food-101}~\cite{bossard14}, 
\textit{Kvasir-v2}~\cite{pogorelov2017kvasir},
\textit{Cars} \cite{krause20133d},
\textit{Intel Images}~\cite{bansal2019intel},
\textit{EuroSAT}~\cite{helber2019eurosat}, 
\textit{Weather} \cite{xiao2021classification},
\textit{Cats and dogs}~\cite{parkhi2012cats},
\textit{MangoLeafBD} \cite{ahmed2023mangoleafbd},
\textit{Beans} \cite{beansdata},
\textit{ImageNet-1K}~\cite{deng2009imagenet},
\textit{Office-Home} ~\cite{venkateswara2017deep},
\textit{PACS}~\cite{li2017deeper}, 
\textit{VLCS}~\cite{Fang_2013_ICCV}.
For more detailed information about the datasets used in the experiments, please refer to Appendix C.1.

\textbf{Models to be merged.} We choose the frequently-used ViT-B/16~\cite{dosovitskiy2020image} throughout our experiments. The model has 12 transformer layers with a dimension of 768. Unless otherwise stated, all its various model weights are downloaded from huggingface's transformer library~\cite{wolf2019huggingface} and are fine-tuned from the ViT-B/16 pretrained on ImageNet-21K~\cite{deng2009imagenet}.

\begin{table}[t]
\vspace{-0.3cm}
\centering
\caption{Performance comparisons when merging ViT models trained on \textbf{large-scale and small-scale datasets}.}
\label{tab:merge_3_models_imnet}
\scalebox{0.62}{
\begin{tabular}{@{}lcccccc@{}}
\toprule
 & \multicolumn{6}{c}{\textbf{Merging Models from large- and small-scale datasets}}\\
\cmidrule(r){1-1} \cmidrule(l){2-7}
Method &Params(M) &  FLOPs(G)&
ImageNet-1K & Cars & EuroSAT & \textbf{Avg($\uparrow$)}\\

\cmidrule(r){1-1} \cmidrule(l){2-7}

Individual & 258.40 & 35.14&80.31 & 85.95 & 98.95 & 88.40 \\
\midrule
AvgMean & 86.75 & 35.14&78.32 & 49.17 & 85.68 & 71.06 \\

\midrule

RegMean & 86.75 & 35.14&78.87 &	78.90 	&97.88 &	85.22  \\

\midrule
Task Arithmetic
&
86.75 & 35.14&75.74 & 83.37 & 95.98 & 85.03 \\
\midrule
Ours + AvgMean \\
$m = 0$& 90.62&35.78&79.58 &	49.61 &	87.52 &	72.24 \\
$m = 1$& 111.88&35.78&79.81 & 61.41 & 90.93 & 77.38 \\
$m = 3$& 154.41&35.78&79.72 & 75.57 & 95.25 & 83.51 \\
$m = 5$& 196.93&35.78&80.02 & 81.80 & 97.71 & 86.51 \\

\midrule
Ours + RegMean \\
$m = 0$& 90.62&35.78&79.27 &	79.21 &	98.07 &	85.52 \\
$m = 1$& 111.88&35.78&79.50 & 80.46 & 98.24 & 86.07 \\
$m = 3$& 154.41&35.78&79.84 & 82.50 & 98.52 & 86.95 \\
$m = 5$& 196.93&35.78&80.04 & 84.19 & 98.69 & 87.64 \\

\bottomrule
\end{tabular}
}
\end{table}

\begin{table*}[t]
\centering
\caption{Domain generalization performance comparisons when merging ViT models fine-tuned on \textbf{different domains}.}
\label{tab:domain_generalization}
\vspace{-0.2cm}
\scalebox{0.51}{
\begin{tabular}{@{}lccccccccccccccccccccc@{}}
\toprule
 & \multicolumn{7}{c}{\textbf{Office-Home}} & \multicolumn{7}{c}{\textbf{PACS}}  & \multicolumn{7}{c}{\textbf{VLCS}} \\
\cmidrule(r){1-1} \cmidrule(lr){2-8} \cmidrule(lr){9-15} \cmidrule(l){16-22}
Method &Params(M)&FLOPs(G)&
\begin{tabular}[c]{@{}c@{}} C,P,R \\ $\downarrow$ \\A\end{tabular} &
\begin{tabular}[c]{@{}c@{}} A,P,R \\ $\downarrow$ \\C\end{tabular} &
\begin{tabular}[c]{@{}c@{}} A,C,R \\ $\downarrow$ \\P\end{tabular} &
\begin{tabular}[c]{@{}c@{}} A,C,P \\ $\downarrow$ \\R\end{tabular} & \textbf{Avg($\uparrow$)} & Params(M)&FLOPs(G)&
\begin{tabular}[c]{@{}c@{}} C,P,S \\ $\downarrow$ \\A\end{tabular} &
\begin{tabular}[c]{@{}c@{}} A,P,S \\ $\downarrow$ \\C\end{tabular} &
\begin{tabular}[c]{@{}c@{}} A,C,S \\ $\downarrow$ \\P\end{tabular} & 
\begin{tabular}[c]{@{}c@{}} A,C,P \\ $\downarrow$ \\S\end{tabular} & \textbf{Avg($\uparrow$)} &Params(M)&FLOPs(G)&
\begin{tabular}[c]{@{}c@{}} L,S,V \\ $\downarrow$ \\C\end{tabular} & 
\begin{tabular}[c]{@{}c@{}} C,S,V \\ $\downarrow$ \\L\end{tabular} & 
\begin{tabular}[c]{@{}c@{}} C,L,V \\ $\downarrow$ \\S\end{tabular} & 
\begin{tabular}[c]{@{}c@{}} C,L,S \\ $\downarrow$ \\V\end{tabular} & \textbf{Avg($\uparrow$)} \\


\cmidrule(r){1-1} \cmidrule(lr){2-8} \cmidrule(lr){9-15} \cmidrule(l){16-22}

AvgMean &
85.97&35.14 &60.98 &37.14 &56.68 &64.27 &54.77 &
 85.84&35.14& 31.88& 36.09 & 38.92 & 41.43 &37.08 &
 85.84&35.14 & 78.94 & 54.40 & 58.90 & 56.78 & 62.26
\\

\cmidrule(r){1-8} \cmidrule(lr){9-15} \cmidrule(l){16-22}

RegMean &
85.97&35.14 &74.99&40.43 &64.59 & 69.20&62.30 &
85.84&35.14& 49.27& 53.54 & 73.77 & 47.08 &55.92 &
85.84&35.14 & 84.02&60.58&65.39 &63.98 & 68.49\\
 
\cmidrule(r){1-8} \cmidrule(lr){9-15} \cmidrule(l){16-22}

Task Arithmatic

&85.97&35.14 &34.03 & 28.91 & 43.55 & 44.32 & 37.70&
85.84&35.14&46.04& 48.72& 63.17 & 49.46 &51.84 &
85.84&35.14 & 87.77 & 62.73 & 68.00 & 66.71 &71.30 \\



\cmidrule(r){1-8} \cmidrule(lr){9-15} \cmidrule(l){16-22}

Ours + AvgMean\\
$m = 0$&
89.84 & 35.78 & 62.05 & 37.87 & 57.13 & 64.56& 55.40&
89.71 & 35.78& 32.42 & 36.69 & 38.80 & 43.22 & 37.78&
89.71 & 35.78& 79.15 & 55.04 &59.14 & 57.56 & 62.72
\\

$m = 1$&
111.11 & 35.78 & 78.66 & 47.93 & 68.26 & 71.93 & 66.70&
110.97& 35.78 &40.97 & 45.73 & 56.29 & 46.96 & 47.49 &
110.97 & 35.78 &82.12 & 58.85 & 61.88 & 63.54 &66.60
\\

$m = 3$  &
153.63& 35.78 &84.54&53.53 & 74.79 & 76.59&72.31&
153.49 & 35.78 &49.37 & 55.33 & 71.01 & 51.13 & 56.71&
153.49 & 35.78 &86.07 & 62.16 & 66.12 & 66.53 & 70.22
\\

$m = 5$ &
196.15 & 35.78& 85.95 & 53.35 & 76.87 & 78.11 & 73.57&
196.02 & 35.78 &53.17 & 58.53 & 76.53 & 50.16 & 59.60&
196.01 & 35.78 & 86.43 & 63.89 & 66.45 & 67.45 & 71.06
\\

\cmidrule(r){1-8} \cmidrule(lr){9-15} \cmidrule(l){16-22}

Ours + RegMean\\
$m = 0 $& 
89.84 & 35.78 &77.21 &	42.58 &	69.16 &	71.79 &	65.19 &
89.71 & 35.78& 49.80 &	53.97 &	76.89 &	48.36 &	57.26 &
89.71 & 35.78&85.37 	&62.01& 	66.27 &	65.61 &	69.82 
\\

$m = 1$ &
111.11 & 35.78&81.21 	&49.28 &	73.06 &	75.53 &	69.77 &
110.97 &35.78&52.59 	&56.61 &77.90 	&48.05& 	58.79& 
110.97 & 35.78&86.43 	&62.54 &	67.09 &	67.45 &	70.88 
\\

$m = 3$ &
153.63& 35.78&84.71 &	54.59 &	76.39 &	77.99 &	73.42& 
153.49& 35.78&56.35 &	58.75 &	84.25 &	50.97 &	62.58 &
153.49 &35.78&87.28 &	62.84 &	68.22 &	68.66 &	71.75 
\\

$m = 5$ & 
196.15 & 35.78&84.96 &	54.36 &	77.61 &	78.86 &	73.95& 
196.02 & 35.78&57.42 &	60.15 &	85.57 &	52.20 &	63.84 &
196.01 & 35.78&87.63 &	63.59 &	68.16 &	68.93 &	72.08 
\\

\bottomrule
\end{tabular}
}
\end{table*}

\begin{table}[t]
\vspace{-0.3cm}
\caption{Merging models adapted to multiple domains on Office-Home dataset. We take domain Products as the source domain and use TVT~\cite{xu2023transferable} to adapt to other domains before merging.}
\label{tab:domain_adaptation}

\centering
\scalebox{0.63}{
\begin{tabular}{@{}lccccccc@{}}
\toprule
 & \multicolumn{7}{c}{\textbf{Office-Home}}\\
\cmidrule(r){1-1}\cmidrule(l){2-8} 
Method &Params(M) &  FLOPs(G)&
A & C & P & R & \textbf{Avg($\uparrow$)}\\

\cmidrule(r){1-1}\cmidrule(l){2-8} 
Individual\\
P $\rightarrow$ A &85.87 & 35.14 & 78.95 & 65.06 & 95.66 & 88.69 & 82.09 \\
P $\rightarrow$ C &85.87 & 35.14 & 75.00 & 74.47 & 98.34 & 87.00 & 83.70\\
P $\rightarrow$ R &85.87 & 35.14 & 78.58 & 64.67 & 95.68 & 88.95 & 81.97\\
\midrule
AvgMean &85.97 & 35.14& 79.20 & 71.49 & 97.12 & 89.17 & 84.25\\
\midrule
RegMean &85.97 & 35.14& 79.32 & 72.55 & 96.94 & 88.96 & 84.44\\
\midrule
Task Arithmetic


&85.97 & 35.14 & 79.42 &  71.21 & 97.10 & 89.24 & 84.24 \\


\midrule
Ours + RegMean\\
$m=0$&89.84 & 35.78&79.55 & 73.70 & 97.21 & 89.19 & 84.91 \\
$m=1$&111.11 & 35.78&79.24 & 73.95 & 97.48 & 89.38 & 85.01 \\
$m=3$ & 153.63& 35.78&79.73 & 73.95 & 97.82 & 89.33 & 85.21 \\
$m=5$ &196.15 & 35.78 & 79.55 & 74.18 & 97.91 & 89.38 & 85.26\\

\bottomrule
\end{tabular}
}
\vspace{-0.2cm}
\end{table}

\textbf{Model merging methods.} We compare the performance of the proposed method with the previously mentioned AvgMean, RegMean, and Task Arithmetic. It should be noted that we validate the performance of Task Arithmetic under the settings of different selection of $\lambda$. We present the best performance under these settings in this section and the entire results is shown in Appendix C.2.
Our method is combined with AvgMean and RegMean, and the performance when selecting different values of $m$ is validated.

\textbf{Gating network.} The gating network is a MobileNetV2 fine-tuned from a pretrained model on ImageNet-1K using no more than 15\% of the test dataset randomly selected from each task or domain. More discussions about the selection of the gating network are shown in \cref{sec:ablation_study}.

\subsection{Merging models with different tasks}

The performance of the merged model is evaluated by per-task accuracy. We compare the performance of different merging methods on merging 3, 6, and 12 models, as shown in Tab.~\ref{tab:merge_3_and_6_models} and~\ref{tab:merge_12_models}. We validate the significance of pretraining in \cref{tab:merge_3_and_6_models}. Thus, in the subsequent experiments, models are all fine-tuned from the pretrained ViT.
To further verify the effectiveness of our method when merging models fine-tuned on datasets ranging from large-scale to small-scale, we merge another 3 models fine-tuned on different tasks including ImageNet-1K, which is a rather challenging task, as shown in Tab.~\ref{tab:merge_3_models_imnet}.
Note that for original AvgMean, RegMean, and Task arithmetic, the classifiers are manually selected, while they can be selected by the gating network if our method is applied.

The results in Tab.~\ref{tab:merge_3_and_6_models}, \ref{tab:merge_12_models}, and \ref{tab:merge_3_models_imnet} show some consistency: after combining our method, the performance of both RegMean and AvgMean is improved. As $m$ increases, the performance of the merged model improves.
Besides, as the number of models increases, it becomes increasingly challenging to merge models, which is reflected in the increasingly larger gaps between the performance of AvgMean and that of individual models on their tasks.
Similarly, the addition of models fine-tuned on more challenging datasets (e.g. ImageNet-1K) increases the difficulty of model merging.
As the difficulty of the merging task increases, existing methods show varying degrees of performance decay. Take merging 12 models as an example, even RegMean, which performs best among existing methods, exhibits a performance decay of nearly 10\%. After combining our method, this value can be reduced to 1.39\% depending on the selection of $m=5$. Although $m=5$ is the setting with the largest number of parameters in the experiment, the parameter number decreases by approximately 50\% compared with 12 individual models. In the case of strict restrictions on the number of parameters of the model, the value of m can be flexibly set to 0 or 1, which can also improve the merged model performance with only a few additional parameters. Moreover, the FLOPs of our method are independent of $m$ and only slightly increase compared with existing methods.

We plot the average task accuracy as $m$ changes when merging 12 models on different tasks in Appendix C.3. To avoid adding extra parameters, we set the stopping point at $m=5$, where the performance of the merged single model is already competitive with $N$ individual models.

\subsection{Merging models with different domains}
\label{sec:exp_domain}
We next shift to different setups where individual models are fine-tuned on different domains of the same task, namely domain generalization and unsupervised domain adaptation setups.

\textbf{Domain generalization.} Our setting is close to leave-one-domain-out cross-validation~\cite{gulrajani2020search}. For a dataset with $M$ domains, each domain is selected as the target domain in turn to validate the domain generalization ability of the models. We merge the $M-1$ models fine-tuned on the source domains and evaluate their performance on the target domain. The result is shown in~\cref{tab:domain_generalization}. 
Compared to AvgMean, RegMean and Task Arithmetic improve the domain generalization ability. Moreover, the performance can be further improved through our method. Similar to the case of merging models for different tasks, the performance is positively related to $m$.
For convenience of comparison, the domain generalization performance of each individual model is also evaluated and shown in Appendix C.4.

\textbf{Unsupervised domain adaptation (UDA).} Under UDA settings where no labels are available in the target domain, some algorithms, e.g. SSRT~\cite{sun2022safe} and TVT~\cite{xu2023transferable}, can be used firstly to obtain models transferred from the source domain to each target domain using only labels from a single source domain. Then, a model applicable to all the target domains can be obtained through model merging. We merge TVT-based models pretrained on ImageNet and fine-tuned on Office-Home. We select the Products domain as the source domain and transfer it to the other three domains. Then we merge the three individual models and the results are shown in \cref{tab:domain_adaptation}. The merged model using the proposed method performs better than other merging methods or any individual model on all the domains, showing the feasibility of combining the proposed method with UDA, thus obtaining a model performing well on all the domains. When taking other domains as the source domain, the results show consistency. Please check Appendix C.5 for more information.

\begin{figure}[t]
  \centering
   \includegraphics[width=1.0\linewidth]{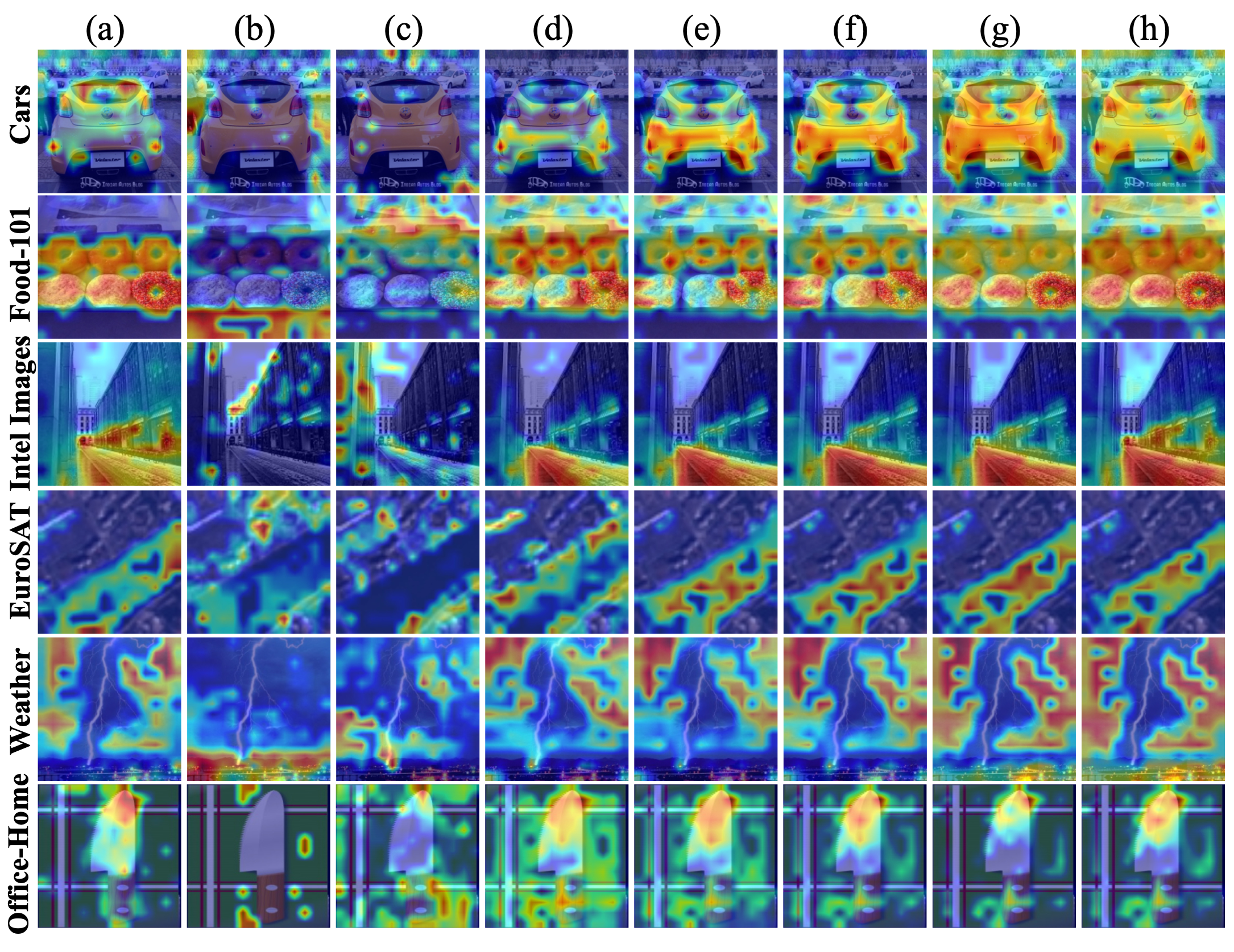}
   \vspace{-6mm}
   \caption{Class activation maps (CAM) of different networks, respectively (a) the 
   model corresponding to the task or domain from which the input image comes, (b) a model not corresponding to the task or the domain, (c) merged model using AvgMean, (d) RegMean, (e) Ours + RegMean (m=0), (f) Ours + RegMean (m=1), (g) Ours + RegMean (m=3), (h) Ours + RegMean (m=5)}
   \label{fig:CAM}
   \vspace{-1mm}
\end{figure}


\subsection{Attention Visualization}
\label{sec:attention_visualization}
We visualize the class activation maps (CAM) of different networks using Grad-CAM~\cite{selvaraju2017grad}.
As shown in \cref{fig:CAM}, our method captures more accurate regions than RegMean and AvgMean. As $m$ increases, our method captures more accurate regions, closer to the individual model corresponding to the task (domain).

\subsection{Ablation study}
\label{sec:ablation_study}
\begin{table}[t]
\caption{The effect of the selection of the gating network architecture on the merging results when merging 12 models on different tasks using RegMean + Ours (m=3).}
\label{tab:ablation_study_gating}
\vspace{-0.2cm}
\centering
\scalebox{0.6}{
\begin{tabular}{@{}ccccc@{}}
\toprule

\begin{tabular}[c]{@{}c@{}} 
Gating Network\\Architecture
\end{tabular}&
\begin{tabular}[c]{@{}c@{}} 
CLIP \\ Fine-tuned
\end{tabular}&
\begin{tabular}[c]{@{}c@{}} 
ResNet-50 \\ Fine-tuned
\end{tabular}&
\begin{tabular}[c]{@{}c@{}} 
MobileNet V2\\From-scratch
\end{tabular}&
\begin{tabular}[c]{@{}c@{}} 
\textbf{MobileNet V2}\\\textbf{Fine-tuned}
\end{tabular}
\\
\cmidrule(r){1-1} \cmidrule(l){2-5}
Acc& 93.83  & 92.52 &17.79&  93.40 
\\
Params(M) & 653.02& 371.76 & 352.10 & 352.10 
\\
FLOPs(G) & 190.70 & 70.28& 35.78 & 35.78 
\\
\bottomrule


\vspace{-4mm}

\end{tabular}
}
\end{table}
\begin{table}[t]
\caption{The effect of weight similarity calculation strategy on the merging results when merging 12 models on different tasks using RegMean + Ours (m=3).}
\label{tab:ablation_study_sim}
\vspace{-0.2cm}
\centering
\scalebox{0.58}{
\begin{tabular}{@{}ccccc@{}}
\toprule

\begin{tabular}[c]{@{}c@{}} 
Calculating\\Strategy
\end{tabular}&
\begin{tabular}[c]{@{}c@{}} 
Seperate-layers +\\Seperate-FFN
\end{tabular}&
\begin{tabular}[c]{@{}c@{}} 
Seperate-layers +\\Combined-FFN
\end{tabular}&
\begin{tabular}[c]{@{}c@{}} 
Concat-layers +\\Seperate-FFN
\end{tabular}&
\begin{tabular}[c]{@{}c@{}} 
\textbf{Concat-layers} \textbf{+}\\\textbf{Combined-FFN}
\end{tabular}

\\
\cmidrule(r){1-1} \cmidrule(l){2-5}
Acc&92.34 & 93.10 & 92.33 & \textbf{93.40}
\\
\bottomrule
\end{tabular}
}
\vspace{-2mm}
\end{table}

\textbf{Architecture and initialization of the gating network.} 
Since the fusion process is performed by the gating network, the architecture and initialization of the gating network may have a significant impact on the merged model's performance. Moreover, the selection of the gating model's architecture determines the FLOPs of the merged model and also decides the number of parameters of the merged model together with $m$.

To investigate the impact of the gating network's architecture and initialization on the merging results, three other models, namely a CLIP~\cite{radford2021learning} initialized ViT-B/16 model, a ResNet-50~\cite{he2016deep} pretrained on ImageNet-1K, and a randomly initialized MobileNet V2, are utilized to merge the 12 models with RegMean and $m = 3$. All the gating networks have gone through the same number of rounds of training or fine-tuning before application. The results are shown in~\cref{tab:ablation_study_gating}. We can see that pretraining-initialization is beneficial to the merging result. We also observe that the impact of the model architecture on the number of parameters and FLOPs is much higher than the impact on accuracy. Compared with a larger gating network, a larger $m$ can more significantly improve the merging performance. Due to comprehensive considerations of performance, FLOPs, and parameter number of the merged model, we select a pretrained MobileNet-V2 in our experiments.

\textbf{Selection of the weight similarity calculation strategy.}
Calculating weight similarity is an important process of our method when $m \neq 0$. Section~\ref{sec:methods} has introduced the method for calculating the weight similarity of a single layer while both attention and MLP layers are composed of more than one layer. Further, four different calculation strategies are selected and compared in~\cref{tab:ablation_study_sim}.

\textit{Concat-layers} strategy concatenates all the related layers before calculating similarity. \textit{Separate-layers} strategy calculates the sum of weight similarity of each related layer. \textit{Combined-MLP} strategy treats two MLPs in an MLP block as a whole while \textit{Separate-MLP} strategy treats two MLPs in an MLP block separately. 
Through the ablation study, \textit{Concat-layers + Combined-MLP} achieve the best and is employed in the proposed method.
\section{Conclusion}
\label{sec:conclusion}
In this paper, we investigate how to merge ViTs from different tasks (domains) for the first time without access to labeled datasets.
We first experimentally show the problems with prior works, then propose a model merging method based on a gating network, through which all kinds of layers can be merged and the classifier can be selected. Further, we propose a weights similarity metric to realize controllable and combined weights merging and boost the performance of the merged model.
The effectiveness of our method is validated by comprehensive experiments on various newly-established benchmarks.
We hope our method can promote the application of model merging.



{
    \small
    \bibliographystyle{ieeenat_fullname}
    \bibliography{main}
}
\appendix
\clearpage
\setcounter{page}{1}
\maketitlesupplementary

\section{Existing model merging methods}
\label{sec:reg and ta}

\textbf{Simple Average Mean (AvgMean)} element-wise averages the weights of all the models to be merged. Though simple, it has been proved effective when merging model weights from different epochs during the training process or merging models fine-tuned on different tasks (domains) from the same initialization \cite{rame2023model, wortsman2022model,sung2023empirical}. Further, AvgMean can be extended as an interpolation by modifying the arithmetic average to a weighted average with given ratios. 
And~\cite{sung2023empirical} finds that a specific merged model as competitive as RegMean can be obtained by adjusting the ratio between models on different modalities. However, for models on different tasks, it is not feasible to obtain such a ratio with the best performance, because the conclusion that vision weight is more important 
than language weight when merging multi-modal models cannot be applied to models on different tasks (domains) but the same modality. Therefore, in this paper, AvgMean arithmetically averages the weights of all the models from different tasks (domains).

\textbf{Task Arithmetic} edits models using task vector, which denotes the difference between the specifically fine-tuned weights and the initial weights~\cite{ilharco2022editing}. For a fine-tuned model weights $W_i$ from a base model weights $W_b$ on a specific task $T_i$, the task vector is defined as $\tau_i=W_i-W_B$. By adding task vectors together, a multi-task model can be built~\cite{ilharco2022editing}, and the merged model weight is $W_M=W_B+\lambda\textstyle\sum_{i=1}^{N}\tau_i$, where $\lambda$ is the scaling parameter, on which the performance of the merged model highly depends. For a specific merging task, once an appropriate $\lambda$ is selected, a high-performance multi-task, multi-domain, or multi-modal model can be produced by Task Arithmetic~\cite{ilharco2022editing, sung2023empirical}.

\textbf{RegMean} merges linear layers by finding closed-form solutions for different weights from the same initialization~\cite{jin2022dataless}. Specifically, when merging $K$ linear model weights $W_i$, where $f_i\left(x\right)=W_i^T x$, $i = 1..K$, the merging problem can be formulated as: \(\min\limits_W\textstyle\sum_{i=1}^{K}\lVert W^TX_i-W_i^TX_i\rVert^2\), where $W$ is the merged model weights, and $X_i$ denotes the input of $i^{th}$ model. The closed-form solution to the problem is: \(W=(\textstyle\sum_{i=1}^{K}X_i^TX_i)^{-1}(\textstyle\sum_{i=1}^{K}X_i^TX_iW_i)\). Moreover, a fixed scale $\alpha$ is introduced as a hyper-parameter for regularization. Specifically, all non-diagonal items of inner product matrices, also known as gram matrices $G_i=X_i^TX_i$, are decreased by multiplying $\alpha$ $(\alpha<1)$. When applied for merging ViTs in this paper, $\alpha$ is set to 0.9, consistent with the setting in \cite{jin2022dataless}.

\begin{table*}[t]
\centering
\caption{Performance comparisons when merging ViT models from 3, 6, and 12 tasks.}
\label{tab:merge_3_6_12_models_appendix}
\vspace{-0.2cm}
\scalebox{0.56}{
\begin{tabular}{@{}lcccccccccccccccc@{}}
\toprule
 & \multicolumn{6}{c}{\textbf{Merging Models from 3 Different Tasks}} & \multicolumn{9}{c}{\textbf{Merging Models from 6 Different Tasks}} \\
\cmidrule(r){1-1} \cmidrule(lr){2-7} \cmidrule(l){8-16} 
Method &Params(M)&FLOPs(G)&MNIST&CIFAR-10 & EuroSAT & \textbf{Avg($\uparrow$)} & Params(M) & FLOPs&MNIST & CIFAR-10 & Vegetables & Food-101 &Kvasir-V2 & EuroSAT&\textbf{Avg($\uparrow$)} \\
\cmidrule(r){1-7} \cmidrule(l){8-16} 
Individual & 257.50& 35.14& 99.22 & 97.88 & 98.95 & 98.65 &515.07&35.14&  
99.22 & 97.88 & 100.00& 87.89 & 93.59&98.95& 96.26 \\

\cmidrule(r){1-7} \cmidrule(l){8-16} 
RegMean& 85.85 &35.14& 98.92 & 97.71 & 97.76 & 98.13 & 85.94&35.14& 98.41 & 96.73 & 97.63 & 85.41 & 80.91 & 94.16 & 92.21 \\
\cmidrule(r){1-7} \cmidrule(l){8-16} 
Task Arithmetic\\
$\lambda=1$& 85.85&35.14 &93.07 & 77.19 & 11.20 & 62.15 & 85.94&35.14&0.04  & 0.30  & 0.00  & 0.00  & 0.90  & 0.75  & 0.33 \\

$\lambda=0.75$
& 85.85&35.14 &98.48 & 94.78 & 89.47 & 94.24 &
85.94&35.14&78.52 & 85.41 & 99.67 & 81.30 & 55.50 & 57.78 & 76.36 \\

$\lambda=0.5$ & 85.85&35.14&96.08 & 93.90 & 82.89 & 90.96 & 85.94&35.14&70.16 & 76.20 & 99.06 & 79.06 & 45.61 & 51.69 & 70.30  \\

$\lambda=0.25$ & 85.85 &35.14&70.87 & 77.62 & 55.15 & 67.88  & 85.94&35.14&42.62 & 55.44 & 93.93 & 72.80 & 29.90 & 34.44 & 54.86\\

\cmidrule(r){1-7} \cmidrule(l){8-16} 

Ours + RegMean\\
$m = 0$ &  89.72&35.78 &98.94 & 97.78 & 98.46 & 98.39& 92.15 &35.78 & 98.48 &96.98 & 97.93 & 85.81 & 85.25 & 95.92 & 93.40 \\

$m = 1$ & 110.98&35.78& 98.95 & 97.83 & 98.56 & 98.45  & 134.67&35.78&98.58 & 94.78 & 99.86 & 85.62 & 89.70 & 96.65 & 94.20 \\

$m = 3$  & 153.50&35.78& 99.02 & 97.83 & 98.67 & 98.51 &
219.72&35.78& 98.75 & 97.54 & 99.86 & 86.90 & 90.35 & 97.40 & 95.13 \\

$m = 5 $&  196.02&35.78& 99.09 & 97.86 & 98.78 & 98.58  & 304.76 & 35.78 & 98.95 & 97.80 & 99.90 & 86.98 & 92.57 & 98.13 & 95.72 \\
\midrule

& \multicolumn{15}{c}{\textbf{Merging Models from 12 Different Tasks}}\\
\cmidrule(r){1-1} \cmidrule(l){2-16}
Method &Params(M) &  FLOPs(G)&
MNIST & CIFAR-10 & Vegetables & Food-101 &Kvasir-v2& Cars & Intel Images & EuroSAT & Weather &Cats\&dogs & Mango & beans & \textbf{Avg($\uparrow$)}\\

\midrule
Individual&1030.20&35.14 &99.22 & 97.88 & 100 & 87.89 & 93.59 & 94.88 & 85.95 & 98.95 & 98.19 & 99.92 & 100 & 97.79&96.19\\
\midrule
RegMean &  86.12&35.14&97.69 & 94.69 & 98.13 & 81.73 & 72.22 & 93.65 & 35.24 & 89.47 & 90.75 & 99.04 & 99.27 & 91.57&86.95\\
\midrule
Task Arithmetic\\
$\lambda=1$ &  86.12&35.14&0.31  & 1.50  & 0.07  & 0.00  & 7.27  & 0.10  & 0.00  & 4.67  & 0.16  & 2.90  & 0.03  & 1.15  &1.51\\
$\lambda=0.75$ 
&  86.12&35.14&48.94 & 65.67 & 93.16 & 74.34 & 35.47 & 90.56 & 10.60 & 31.45 & 79.07 & 96.27 & 92.55 & 84.29 &66.86\\
$\lambda=0.5$ &  86.12&35.14&43.64 & 56.35 & 93.70 & 73.12 & 31.10 & 88.73 & 9.77  & 35.05 & 75.11 & 95.71 & 88.37 & 85.06 &64.64\\
$\lambda=0.25$ &  86.12&35.14&30.31 & 43.04 & 83.53 & 68.21 & 25.06 & 82.86 & 7.80  & 24.57 & 62.14 & 92.70 & 72.00 & 70.88&55.26\\
\midrule
Ours + RegMean\\
$m = 0$&96.97&35.78&97.80  & 95.94 & 98.16 & 81.89 & 78.86 & 93.56 & 39.77 & 92.68 & 91.71 & 98.93 & 99.62 & 91.95 &88.41\\

$m = 1$&182.01&35.78&97.95 & 96.04 & 98.26 & 81.75 & 84.15 & 93.73 & 42.89 & 94.09 & 92.90  & 98.96 & 99.80  & 91.95 &89.37\\

$m = 3$ & 352.10 & 35.78&98.65 &97.65 & 98.40 &85.09& 90.38 & 94.70 & 70.49 & 97.09 & 95.76 & 99.11 & 99.95 & 93.49 & 93.40\\

$m = 5$&522.19&35.78&98.98 & 97.97 & 98.53 & 86.22 & 92.52 & 94.40  & 79.95 & 98.06 & 96.88 & 99.15 & 99.95 & 95.02 &94.80\\

\bottomrule
\end{tabular}
}
\end{table*}
\begin{table*}[t]
\centering
\caption{Domain generalization performance comparisons when merging ViT models fine-tuned on different domains.}
\label{tab:TA_domain_generalization_appendix}
\vspace{-0.2cm}
\scalebox{0.51}{
\begin{tabular}{@{}lccccccccccccccccccccc@{}}
\toprule
 & \multicolumn{7}{c}{\textbf{Office-Home}} & \multicolumn{7}{c}{\textbf{PACS}}  & \multicolumn{7}{c}{\textbf{VLCS}} \\
\cmidrule(r){1-1} \cmidrule(lr){2-8} \cmidrule(lr){9-15} \cmidrule(l){16-22}
Method &Params(M)&FLOPs(G)&
\begin{tabular}[c]{@{}c@{}} C,P,R \\ $\downarrow$ \\A\end{tabular} &
\begin{tabular}[c]{@{}c@{}} A,P,R \\ $\downarrow$ \\C\end{tabular} &
\begin{tabular}[c]{@{}c@{}} A,C,R \\ $\downarrow$ \\P\end{tabular} &
\begin{tabular}[c]{@{}c@{}} A,C,P \\ $\downarrow$ \\R\end{tabular} & \textbf{Avg($\uparrow$)} & Params(M)&FLOPs(G)&
\begin{tabular}[c]{@{}c@{}} C,P,S \\ $\downarrow$ \\A\end{tabular} &
\begin{tabular}[c]{@{}c@{}} A,P,S \\ $\downarrow$ \\C\end{tabular} &
\begin{tabular}[c]{@{}c@{}} A,C,S \\ $\downarrow$ \\P\end{tabular} & 
\begin{tabular}[c]{@{}c@{}} A,C,P \\ $\downarrow$ \\S\end{tabular} & \textbf{Avg($\uparrow$)} &Params(M)&FLOPs(G)&
\begin{tabular}[c]{@{}c@{}} L,S,V \\ $\downarrow$ \\C\end{tabular} & 
\begin{tabular}[c]{@{}c@{}} C,S,V \\ $\downarrow$ \\L\end{tabular} & 
\begin{tabular}[c]{@{}c@{}} C,L,V \\ $\downarrow$ \\S\end{tabular} & 
\begin{tabular}[c]{@{}c@{}} C,L,S \\ $\downarrow$ \\V\end{tabular} & \textbf{Avg($\uparrow$)} \\


\cmidrule(r){1-1} \cmidrule(lr){2-8} \cmidrule(lr){9-15} \cmidrule(l){16-22}
RegMean &
85.97&35.14 &74.99&40.43 &64.59 & 69.20&62.30 &
85.84&35.14& 49.27& 53.54 & 73.77 & 47.08 &55.92 &
85.84&35.14 & 84.02&60.58&65.39 &63.98 & 68.49\\
\cmidrule(r){1-8} \cmidrule(lr){9-15} \cmidrule(l){16-22}

Task Arithmatic\\
$\lambda=1$&
85.97&35.14 &8.41& 7.10&11.47 &11.15 & 9.53&
85.84&35.14& 39.63 & 53.79 & 61.81 & 48.48 &50.93 &
85.84&35.14 & 85.99 & 59.82 & 65.70 & 66.44 &69.49 \\

$\lambda=0.75$ 
&85.97&35.14 &34.03 & 28.91 & 43.55 & 44.32 & 37.70&
85.84&35.14&46.04& 48.72& 63.17 & 49.46 &51.84 &
85.84&35.14 & 87.77 & 62.73 & 68.00 & 66.71 &71.30 \\

$\lambda=0.5$  &
85.97&35.14&62.13&37.14 & 56.68& 64.27 & 50.06&
85.84&35.14& 31.88 & 36.09 & 38.92 & 43.22 & 37.53&
85.84&35.14 & 78.94 & 54.40 & 58.90 & 56.78 & 62.26\\

$\lambda=0.25$ &
85.97&35.14 &7.62&10.42 & 10.95& 36.10 &16.27 &
85.84&35.14&16.46 & 16.42 & 18.98 & 29.68 & 20.39&
85.84&35.14 & 60.42 & 28.50 & 41.10 & 35.54 & 41.39\\

\cmidrule(r){1-8} \cmidrule(lr){9-15} \cmidrule(l){16-22}

Ours + RegMean\\
$m = 0 $& 
89.84 & 35.78 &77.21 &	42.58 &	69.16 &	71.79 &	65.19 &
89.71 & 35.78& 49.80 &	53.97 &	76.89 &	48.36 &	57.26 &
89.71 & 35.78&85.37 	&62.01& 	66.27 &	65.61 &	69.82 
\\

$m = 1$ &
111.11 & 35.78&81.21 	&49.28 &	73.06 &	75.53 &	69.77 &
110.97 &35.78&52.59 	&56.61 &77.90 	&48.05& 	58.79& 
110.97 & 35.78&86.43 	&62.54 &	67.09 &	67.45 &	70.88 
\\

$m = 3$ &
153.63& 35.78&84.71 &	54.59 &	76.39 &	77.99 &	73.42& 
153.49& 35.78&56.35 &	58.75 &	84.25 &	50.97 &	62.58 &
153.49 &35.78&87.28 &	62.84 &	68.22 &	68.66 &	71.75 
\\

$m = 5$ & 
196.15 & 35.78&84.96 &	54.36 &	77.61 &	78.86 &	73.95& 
196.02 & 35.78&57.42 &	60.15 &	85.57 &	52.20 &	63.84 &
196.01 & 35.78&87.63 &	63.59 &	68.16 &	68.93 &	72.08 
\\

\bottomrule
\end{tabular}
}
\end{table*}
\begin{table}[t]
\vspace{-0.3cm}
\centering
\caption{Performance comparisons when merging ViT models trained on large-scale and small-scale datasets}
\label{tab:merge_3_models_imnet_appendix}
\scalebox{0.62}{
\begin{tabular}{@{}lcccccc@{}}
\toprule
 & \multicolumn{6}{c}{\textbf{Merging Models from large- and small-scale datasets}}\\
\cmidrule(r){1-1} \cmidrule(l){2-7}
Method &Params(M) &  FLOPs(G)&
ImageNet-1K & Cars & EuroSAT & \textbf{Avg($\uparrow$)}\\

\cmidrule(r){1-1} \cmidrule(l){2-7}

Individual & 258.40 & 35.14&80.31 & 85.95 & 98.95 & 88.40 \\

\midrule

RegMean & 86.75 & 35.14&78.87 &	78.90 	&97.88 &	85.22  \\

\midrule
Task Arithmetic\\
$\lambda=1$ & 86.75 & 35.14&75.74 & 83.37 & 95.98 & 85.03 \\
$\lambda=0.75$& 86.75 & 35.14&77.92 & 74.29 & 93.74 & 81.98 \\
$\lambda=0.5$ & 86.75 & 35.14&79.32 & 49.17 & 85.68 & 71.39 \\
$\lambda=0.25$ & 86.75 & 35.14&80.07 & 17.36 & 57.24 & 51.56\\

\midrule
Ours + RegMean \\
$m = 0$& 90.62&35.78&79.27 &	79.21 &	98.07 &	85.52 \\
$m = 1$& 111.88&35.78&79.50 & 80.46 & 98.24 & 86.07 \\
$m = 3$& 154.41&35.78&79.84 & 82.50 & 98.52 & 86.95 \\
$m = 5$& 196.93&35.78&80.04 & 84.19 & 98.69 & 87.64 \\

\bottomrule
\end{tabular}
}
\end{table}

\begin{table}[t]
\vspace{-0.3cm}
\caption{Merging models adapted to multiple domains on Office-Home dataset. We take domain Products as the source domain and adapt to other domains before merging.}
\label{tab:TA_domain_adaptation}

\centering
\scalebox{0.63}{
\begin{tabular}{@{}lccccccc@{}}
\toprule
 & \multicolumn{7}{c}{\textbf{Office-Home}}\\
\cmidrule(r){1-1}\cmidrule(l){2-8} 
Method &Params(M) &  FLOPs(G)&
A & C & P & R & \textbf{Avg($\uparrow$)}\\

\cmidrule(r){1-1}\cmidrule(l){2-8} 
Individual\\
P $\rightarrow$ A &85.87 & 35.14 & 78.95 & 65.06 & 95.66 & 88.69 & 82.09 \\
P $\rightarrow$ C &85.87 & 35.14 & 75.00 & 74.47 & 98.34 & 87.00 & 83.70\\
P $\rightarrow$ R &85.87 & 35.14 & 78.58 & 64.67 & 95.68 & 88.95 & 81.97\\
\midrule
RegMean &85.97 & 35.14& 79.32 & 72.55 & 96.94 & 88.96 & 84.44\\
\midrule
Task Arithmetic\\

$\lambda=1$ &85.97 & 35.14 & 71.27 & 67.79 & 96.80 & 84.79 & 80.16 \\

$\lambda=0.75$ &85.97 & 35.14 & 77.16 & 71.10 & 97.48 & 88.03 & 83.44 \\
$\lambda=0.5$&85.97 & 35.14 & 79.42 &  71.21 & 97.10 & 89.24 & 84.24 \\

$\lambda=0.25$ &85.97 & 35.14 &79.63 & 68.50 & 95.14 & 88.80 & 83.02\\

\midrule
Ours + RegMean\\
$m=0$&89.84 & 35.78&79.55 & 73.70 & 97.21 & 89.19 & 84.91 \\
$m=1$&111.11 & 35.78&79.24 & 73.95 & 97.48 & 89.38 & 85.01 \\
$m=3$ & 153.63& 35.78&79.73 & 73.95 & 97.82 & 89.33 & 85.21 \\
$m=5$ &196.15 & 35.78 & 79.55 & 74.18 & 97.91 & 89.38 & 85.26\\

\bottomrule
\end{tabular}
}
\vspace{-0.2cm}
\end{table}
\begin{table*}[t]
\centering
\caption{Domain generalization performance of individual models.}
\label{tab:domain_inividual}
\vspace{-0.2cm}
\scalebox{0.7}{
\begin{tabular}{@{}cccccccccccccccccc@{}}
\toprule

 & \multicolumn{5}{c}{\textbf{Office-Home}}&&
\multicolumn{5}{c}{\textbf{PACS}}&&
\multicolumn{5}{c}{\textbf{VLCS}}
\\
 
\cmidrule(r){1-1} \cmidrule(lr){2-6}\cmidrule(lr){7-7} \cmidrule(lr){8-12}
\cmidrule(lr){13-13} \cmidrule(l){14-18}

\begin{tabular}[c]{@{}c@{}} 
Target($\rightarrow$) \\ Source($\downarrow$)
\end{tabular}  & A &C &P &R&\textbf{Avg}&
\begin{tabular}[c]{@{}c@{}} 
Target($\rightarrow$) \\ 
Source($\downarrow$)
\end{tabular}  & A &C &P &S&\textbf{Avg}&
\begin{tabular}[c]{@{}c@{}} 
Target($\rightarrow$) \\ Source($\downarrow$)
\end{tabular}  
& C&L &S &V&\textbf{Avg}\\

\cmidrule(r){1-1} \cmidrule(lr){2-6}\cmidrule(lr){7-7} \cmidrule(lr){8-12}
\cmidrule(lr){13-13} \cmidrule(l){14-18}

A&$-$&57.80 & 70.93 & 78.32 & 69.02&
A&$-$&66.53 & 85.93 & 34.92 & 62.46&
C&$-$&44.64 & 60.05 &49.00 & 51.23
\\
C&66.51&$-$&61.66 & 66.68&64.95&
C&73.78 & $-$& 57.66 & 49.10 & 60.18& 
L&85.31&$-$&47.12 &67.67 &66.70
\\
P&71.60 & 50.71 & $-$ & 76.54&66.28&
P&58.15 & 34.80 & $-$ & 24.02 &38.99&
S&66.42 & 63.73 & $-$ & 67.92 & 66.02
\\
R & 85.20 & 48.02 & 78.47 &$-$ &70.56&
S&33.83 & 49.66 & 24.12 & $-$ & 35.87&
V&87.92 & 56.02 & 69.07 & $-$&71.00
\\

\bottomrule

\end{tabular}
}
\end{table*}

\begin{table*}[t]
\caption{Merging models adapted to multiple domains on Office-Home dataset}
\label{tab:domain_adaptation_appendix}
\vspace{-0.2cm}
\centering
\scalebox{0.48}{
\begin{tabular}{@{}lccccccccccccccccccccccc@{}}
\toprule
 & \multicolumn{23}{c}{\textbf{Office-Home}}\\
\cmidrule(r){1-1}\cmidrule(lr){2-8} \cmidrule(lr){9-9} \cmidrule{10-16}\cmidrule(lr){17-17} \cmidrule{18-24}

Method &Params(M) &  FLOPs(G)&
A & C & P & R & \textbf{Avg($\uparrow$)}&Method &Params(M) &  FLOPs(G)&
A & C & P & R & \textbf{Avg($\uparrow$)}&Method &Params(M) &  FLOPs(G)&
A & C & P & R & \textbf{Avg($\uparrow$)}\\

\cmidrule(r){1-1}\cmidrule(lr){2-8} \cmidrule(lr){9-9} \cmidrule{10-16}\cmidrule(lr){17-17} \cmidrule{18-24}

Individual\\
A $\rightarrow$ C &85.87 & 35.14 & 98.56&73.16&83.58 & 86.70 & 85.50&
C $\rightarrow$ A &85.87 & 35.14 & 81.48 & 94.79 & 83.50 & 84.47 & 85.77&
R $\rightarrow$ A&85.87 & 35.14 &84.38 & 65.23 & 88.78 & 96.36 & 83.69
\\

A $\rightarrow$ P &85.87 & 35.14 & 93.22 & 65.63 & 84.46 & 87.43 & 82.69&
C $\rightarrow$ P &85.87 & 35.14 &76.42 & 92.41 & 84.76 & 85.32 & 85.89&
R $\rightarrow$ C &85.87 & 35.14& 79.92 & 74.90 & 88.89 & 98.88 & 85.65 
\\

A $\rightarrow$ R &85.87 & 35.14 & 99.14 & 65.23 & 84.14 & 88.66 & 84.29&
C $\rightarrow$ R &85.87 & 35.14 &76.22 & 95.19 & 84.53 & 85.42 & 85.96& 
R $\rightarrow$ P &85.87 & 35.14&79.73 & 64.17 & 89.79 & 96.31 & 82.50 
\\

\midrule
AvgMean &85.97 & 35.14& 97.82 & 70.97 & 85.23 & 88.19 & 85.55&
& 85.97 & 35.14&79.38 & 94.55 & 83.22 & 85.89 & 86.22&
& 85.97 & 35.14&82.82 & 72.28 & 89.14 & 97.44 & 85.42
\\

\midrule

RegMean &85.97 & 35.14& 97.82 & 72.25 & 84.35 &88.28 & 85.68&
& 85.97 & 35.14&79.53 & 94.34 & 83.50 & 85.60 & 85.74&
& 85.97 & 35.14&83.31 & 73.86 & 89.43 & 97.48 & 86.02
\\

\midrule
Task Arithmetic\\
$\lambda=1$ &85.97 & 35.14 & 93.13 & 62.93 & 72.49 & 80.27 & 77.21&
&85.97 & 35.14 &49.91 & 78.85 & 58.66 & 59.38 & 61.70&
&85.97 & 35.14&72.90 & 65.35 & 83.94 & 95.05 & 79.31
\\

$\lambda=0.75$ &85.97 & 35.14 &97.94 & 69.10 & 81.06 & 86.01 & 83.53 &
&85.97 & 35.14 &71.96 & 93.29 & 77.24 & 78.31 & 80.20&
&85.97 & 35.14 &80.57 & 71.16 & 88.48 & 97.80 & 84.50
\\

$\lambda=0.5$&85.97 & 35.14 & 97.68 & 71.02 & 85.23 & 88.13 & 85.52 &
&85.97 & 35.14 &79.32 & 94.61 & 84.21 & 85.42 & 85.89&
&85.97 & 35.14 &82.70 & 72.04 & 89.77 & 97.41 & 85.48 
\\

$\lambda=0.25$ &85.97 & 35.14 &94.43 & 69.04 & 86.44 & 88.42 & 84.58&
&85.97 & 35.14 &80.55 & 88.71 & 83.69 & 84.39 & 84.34 &
&85.97 & 35.14 &82.05 & 68.59 & 89.23 & 95.83 & 83.93 

\\

\midrule
Ours + RegMean\\
$m=0$&89.84 & 35.78&98.31 & 72.28 & 83.97 & 88.53 & 85.77&
&89.84 & 35.78&80.63 & 94.43 & 83.64 & 85.55 & 86.06&
&89.84 & 35.78&83.86 & 74.46 & 89.52 & 97.60 & 86.36

\\

$m=1$&111.11 & 35.78&98.48 & 72.41 & 84.14 & 88.53 & 85.89&
&111.11 & 35.78&80.58 & 94.55 & 83.50 & 85.71 & 86.09&
&111.11 & 35.78&84.09 & 74.48 & 89.49 & 97.85 & 86.48
\\

$m=3$ & 153.63& 35.78&98.77 & 72.58 & 83.97 & 88.53 & 85.96&
 & 153.63& 35.78& 80.82 & 94.61 & 83.85 & 85.60 & 86.22 &
  & 153.63& 35.78& 83.90 & 74.75 & 89.57 & 98.12 & 86.59 
\\

$m=5$ &196.15 & 35.78 & 98.85 & 72.69 & 84.73 & 88.60 & 86.22&
 &196.15 & 35.78 &81.01 & 94.71 & 83.85 & 85.71 & 86.32&
 &196.15 & 35.78 & 84.09 & 74.57 & 89.55 & 98.37 & 86.65
\\

\bottomrule
\end{tabular}
}
\end{table*}

\section{Algorithm flow of the proposed method}
\label{sec:algo}
The algorithm flow of the proposed method is presented in \cref{alg:alg0}.
We obtain a merged model $f_M$ using the given $N$ models [$f_1..f_N$], where $f_k=[W_k^1...W_k^l, C_k], k \in [1..N]$.
Following previous works~\cite{rame2023model, wortsman2022model,sung2023empirical}, all the models $f_1..f_N$ are from the same initialization and fine-tuned on different tasks (domains) $[T_1...T_N]$.
A gating network $G$ needs to be trained in advance before the merging process using some of the unlabeled datasets [$X_1..X_N$] from the test or validation sets of all the tasks (domains).
The controllable and combined layer merging process is based on weight similarity of each layer calculated by \cref{eq:sim}.

\begin{algorithm}[tb]
   \caption{The proposed merging method}
   \label{alg:alg0}
   
\begin{algorithmic}[1]
   \STATE {\bfseries Data:} Given $N$ models $[f_1...f_N]$ fine-tuned from the same initialization and $N$ corresponding unlabeled datasets $[X_1...X_N]$, where $f=[W^1...W^l, C]$, $l$ and $C$ denote the layer number and classifier respectively.
   \STATE {\bfseries Result:} Merged Model $f_M$ 

\STATE \textcolor{JungleGreen}{$\triangleright$ \textit{Prepare Gating Network and Weight Similarity}}

\STATE Train the gating network $G$ via $[X_1..X_N]$

\STATE Compute the weight similarity of each layer via Eq.~\ref{eq:sim}, and give a threshold $th$ for $[Sim(L^1)...Sim(L^l)]$.

\STATE \textcolor{JungleGreen}{$\triangleright$ \textit{Controllable
and Combined Layer Merging Process}}
\FOR{$j$ in 1,2,...,$l$}
\IF{$Sim(L^j) \leq th$}
    \STATE \textcolor{JungleGreen}{$\triangleright$ \textit{Merge through Gating Given Input $x$}}
    \STATE $P(x) \gets SoftMax(G(x))$
    \STATE $W_M^j \gets \sum_{i=1}^{N} p_i \cdot W_i^j$  
    \ELSE
    \STATE \textcolor{JungleGreen}{$\triangleright$ \textit{Merge through AvgMean or RegMean}}
    \STATE $W_M^j \gets$ AvgMean $\OR$ RegMean for $[W_1^j..W_N^j]$
    \ENDIF
\ENDFOR
\STATE \textcolor{JungleGreen}{$\triangleright$ \textit{Classifier Selection}}
\STATE $C_M \gets C_{argmax( P(X) )}$
\RETURN $f_M \gets [W_M^1..W_M^l, C_M]$

\end{algorithmic}
\end{algorithm}

\section{More details and experiments}

\subsection{Datasets}
\label{sec:datasets}
Our experiments involve 16 different  datasets for finetuning their respective model weights. \textit{MNIST}~\cite{lecun1998mnist}, a 10-class dataset of handwritten digits from 0 to 9. \textit{CIFAR-10}~\cite{krizhevsky2009learning}, a 10-class dataset of some vehicles and animals including cats, dogs, and trucks. \textit{Vegetables} \cite{ahmed2021dcnn}, a 15-class dataset of some common vegetables including carrots, potatoes, and pumpkins. \textit{Food-101}~\cite{bossard14}, a 101-class dataset of food, including donuts, waffles, and chocolate cakes. \textit{Kvasir-v2}~\cite{pogorelov2017kvasir}, an 8-class dataset of pictures from inside the gastrointestinal tract for disease grading. \textit{Cars} \cite{krause20133d}, a 196-class dataset of cars depending on their make, model, and year. \textit{Intel Images}~\cite{bansal2019intel}, a 6-class dataset of natural scenes including buildings, sea, and forests. \textit{EuroSAT}~\cite{helber2019eurosat}, a 10-class dataset for land use and land cover classification. \textit{Weather} \cite{xiao2021classification}, an 11-class dataset for weather phenomena classification including snow, rain, and lightning. \textit{Cats and dogs}~\cite{parkhi2012cats}, a 2-class dataset to distinguish dogs from cats. \textit{MangoLeafBD} \cite{ahmed2023mangoleafbd}, an 8-class (including the healthy category) dataset of mango leaf images to identify 7 mango diseases. \textit{Beans} \cite{beansdata}, a 3-class (including the healthy category) dataset of bean leaf images to identify 2 bean diseases. \textit{ImageNet-1K}~\cite{deng2009imagenet}, a 1000-class dataset for image classification at large scale. \textit{Office-Home} ~\cite{venkateswara2017deep}, a 65-class dataset containing images of a variety of everyday objects from 4 domains. \textit{PACS}~\cite{li2017deeper}, a 7-class dataset covering photo, sketch, cartoon, and painting domains. \textit{VLCS}~\cite{Fang_2013_ICCV}, a 5-class dataset selected from four standard datasets.

\subsection{Performance of Task Arithmetic with different selection of $\lambda$}
\label{sec:ta_lambda}
The performance of Task Arithmetic highly depends on the selection of hyper-parameter $\lambda$. In \cref{sec:results}, we present the best performance under different selections of $\lambda$.
Here, we present the entire experimental results of Task Arithmetic under the settings of $\lambda = 1, 0.75, 0.5, 0.25$, along with the results of RegMean and Ours + RegMean for comparison in Tab.~\ref{tab:merge_3_6_12_models_appendix}, 
\ref{tab:TA_domain_generalization_appendix}, 
\ref{tab:merge_3_models_imnet_appendix}, 
\ref{tab:TA_domain_adaptation},

\subsection{The impact of hyper-parameter selection of \textit{m} on the merged model's performance}

\label{sec:hyper_param_m}
\begin{figure}[t]
  \centering
   \includegraphics[width=1.0\linewidth]{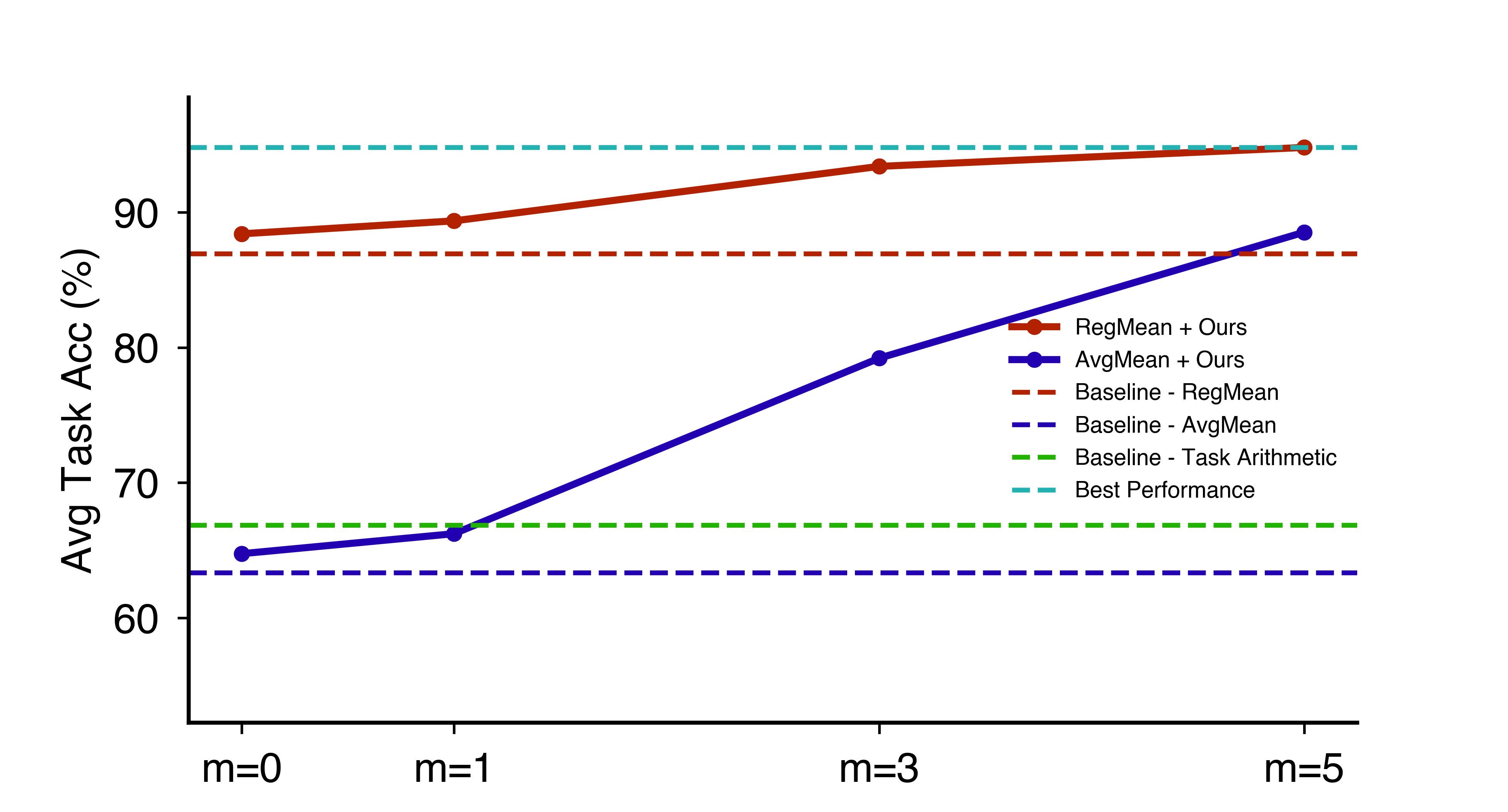}
   \caption{The effect of the selection of hyper-parameter $m$ on the merging results.}
   \label{fig:hyperparam_m}
   \vspace{-0.2cm}
\end{figure}

\cref{fig:hyperparam_m} shows the relationship between the merged model's performance and $m$ when merging 12 models on different tasks.
By increasing $m$, the performance of the merged model significantly improves whether our method is combined with AvgMean or RegMean. In conclusion, this is an effective way to improve the merged model's performance, especially when the merging task is challenging, i.e., there are numerous models to be merged or models on challenging tasks are included.

\begin{figure*}[t]
  \centering
   \includegraphics[width=1.0\linewidth]{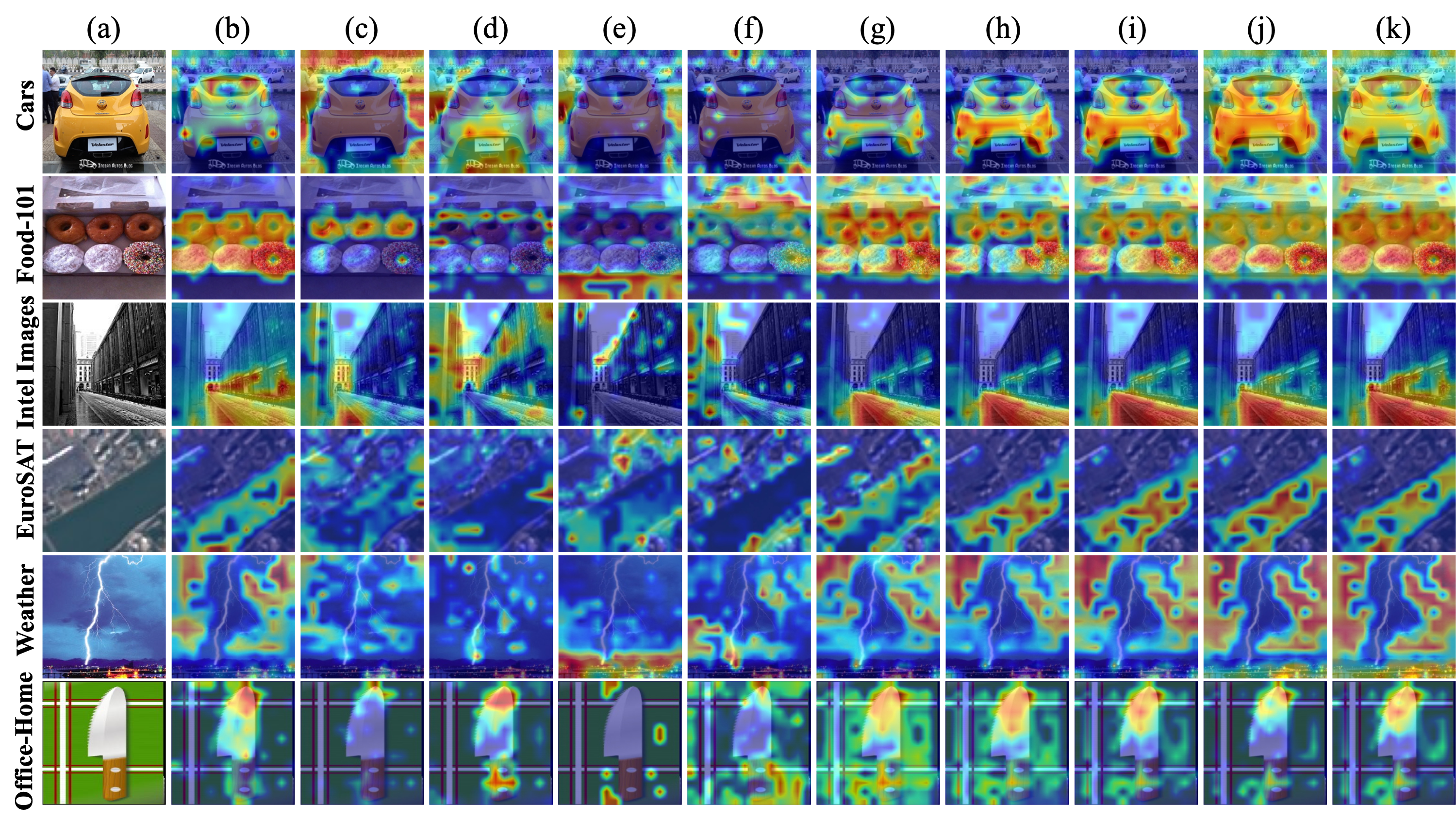}
   \caption{CAMs of different networks, respectively 
   (a) the original image, (b) the model corresponding to the task or domain from which the input image comes, (c-e) models not corresponding to the task or the domain, (f) merged model using AvgMean, (g) RegMean, (h) Ours + RegMean (m=0), (i) Ours + RegMean (m=1), (j) Ours + RegMean (m=3), (k) Ours + RegMean (m=5).}
   \label{fig:heatmap_appendix}
   \vspace{-0.2cm}
\end{figure*}

\subsection{Domain generalization ability of individual models}
\label{sec:single_dg}
In \cref{tab:domain_inividual}, we present the domain generalization ability of individual models used in \cref{sec:exp_domain}. 
Combined with \cref{tab:domain_generalization}, it strongly suggests that the proposed method can improve the domain generalization performance of the merged model compared with each individual model.

\subsection{More results with domain adaptation}
\label{sec:da_more_results}

In~\cref{sec:exp_domain}, we present the merging results when combining the proposed method with TVT~\cite{xu2023transferable}, a UDA method, on the Office-Home dataset. We present the results in~\cref{tab:domain_adaptation} when the Products domain is selected as the source domain, and we adapt the model to the other three domains before merging. 
Here, we present additional experimental results when the other 3 domains are respectively selected as the source domain. The results are shown in \cref{tab:domain_adaptation_appendix}.

\subsection{More attention visualization results}
\label{sec:more_heatmaps}
In \cref{sec:attention_visualization}, we visualize some CAMs of different networks. Here, in \cref{fig:heatmap_appendix}, we present more detailed information, including the original images and more visualization results for models not corresponding to the task (domain).

\section{Limitations and future work}
\label{sec:limitations}
Despite the good quantitative results, the proposed method suffers from several limitations. 
First, the proposed method requires additional computational sources to train the gating network before merging. 
Second, when applied to ViTs trained from scratch instead of fine-tuned from a pretrained model, the proposed method may easily fail. 
Third, the proposed method can only be applied to image classification problems currently. When applied to other vision tasks, including object detection and image segmentation, our method may need some additional adaptations.

We observe that the performance improvement comes with the increment in parameter numbers, depending on the selection of hyper-parameter $m$. We believe that improving the merged models' performance while maintaining the parameter numbers is a significant direction for future work.



\end{document}